\documentclass[twoside]{article}

\usepackage[accepted]{aistats2024}
\usepackage{amsmath,amsfonts,bm}

\def\eqref#1{equation~\ref{#1}}
\def\1{\bm{1}}

\DeclareMathAlphabet{\mathsfit}{\encodingdefault}{\sfdefault}{m}{sl}
\SetMathAlphabet{\mathsfit}{bold}{\encodingdefault}{\sfdefault}{bx}{n}

\usepackage[utf8]{inputenc}
\usepackage[T1]{fontenc}

\usepackage{paralist}            % compact enumerate
\usepackage{xcolor}              % color
\usepackage{mathtools}           % comment under arrow
\usepackage{comment}             % comment out
\usepackage{hyperref}
\usepackage[noabbrev,nameinlink]{cleveref}
\usepackage{pifont}
\usepackage{minitoc}

\usepackage{url}            
\usepackage{graphicx}    
\usepackage{booktabs}  
\usepackage{algorithm, algpseudocode}%
\floatname{algorithm}{Algorithm}

\usepackage{amsthm}

\algtext*{EndFunction}
\usepackage{amsmath}
\usepackage{bm}
\usepackage{amsfonts} 
\usepackage{dsfont}
\usepackage{wrapfig}
\usepackage{graphicx}
\usepackage{natbib}
\usepackage{amsthm}
\usepackage{caption}
\usepackage{subcaption}
\usepackage{caption}
\usepackage{amssymb}
\usepackage{wrapfig}
\usepackage[normalem]{ulem}
\usepackage{nicefrac}       % compact symbols for 1/2, etc.
\usepackage{microtype}      % microtypography
\usepackage{lipsum}
\usepackage{graphicx}
\usepackage{caption}
\usepackage{xcolor}
\usepackage{upgreek}
\usepackage{thmtools}
\usepackage{thm-restate}
\usepackage{amsmath}

\newtheorem{prop}{Proposition}[]

\begin{document}

% If your paper is accepted and the title of your paper is very long,
% the style will print as headings an error message. Use the following
% command to supply a shorter title of your paper so that it can be
% used as headings.
%
%\runningtitle{I use this title instead because the last one was very long}

% If your paper is accepted and the number of authors is large, the
% style will print as headings an error message. Use the following
% command to supply a shorter version of the authors names so that
% they can be used as headings (for example, use only the surnames)
%
%\runningauthor{Surname 1, Surname 2, Surname 3, ...., Surname n}

\twocolumn[
\aistatstitle{Adaptive Batch Sizes for Active Learning:\\
A Probabilistic Numerics Approach}

\aistatsauthor{Masaki Adachi$^{1,3}$ \And Satoshi Hayakawa$^2$ \And Martin Jørgensen$^4$ \And Xingchen Wan$^1$}
\aistatsauthor{Vu Nguyen$^5$ \And Harald Oberhauser$^2$ \And Michael A. Osborne$^1$}
\aistatsaddress{
$^1$Machine Learning Research Group, University of Oxford\\
$^2$Mathematical Institute, University of Oxford\\
$^3$Toyota Motor Corporation\\
$^4$Department of Computer Science, University of Helsinki\\
$^5$Amazon}
]
\runningtitle{Adaptive Batch Sizes for Active Learning}

\begin{abstract}
  Active learning parallelization is widely used, but typically relies on fixing the batch size throughout experimentation. This fixed approach is inefficient because of a dynamic trade-off between cost and speed---larger batches are more costly, smaller batches lead to slower wall-clock run-times---and the trade-off may change over the run (larger batches are often preferable earlier). To address this trade-off, we propose a novel Probabilistic Numerics framework that adaptively changes batch sizes.
  By framing batch selection as a quadrature task, our integration-error-aware algorithm facilitates the automatic tuning of batch sizes to meet predefined quadrature precision objectives, akin to how typical optimizers terminate based on convergence thresholds. This approach obviates the necessity for exhaustive searches across all potential batch sizes.
  %We define batch construction as a \emph{quantization} task---a task of approximating a continuous target distribution (e.g. acquisition function), with a discrete distribution (batch samples). We measure the batch quality through the divergence between these two distributions. Instead of a set batch size, we fix the \emph{precision} of approximation, allowing dynamic adjustment of batch size and query locations.
  We also extend this to scenarios with constrained active learning and constrained optimization, interpreting constraint violations as reductions in the precision requirement, to subsequently adapt batch construction.
  Through extensive experiments, we demonstrate that our approach significantly enhances learning efficiency and flexibility in diverse Bayesian batch active learning and Bayesian optimization applications.
\end{abstract}

\doparttoc % Tell to minitoc to generate a toc for the parts
\faketableofcontents % Run a fake tableofcontents command for the partocs
\section{Introduction}
\begin{figure}
  \centering
  \includegraphics[width=0.65\hsize]{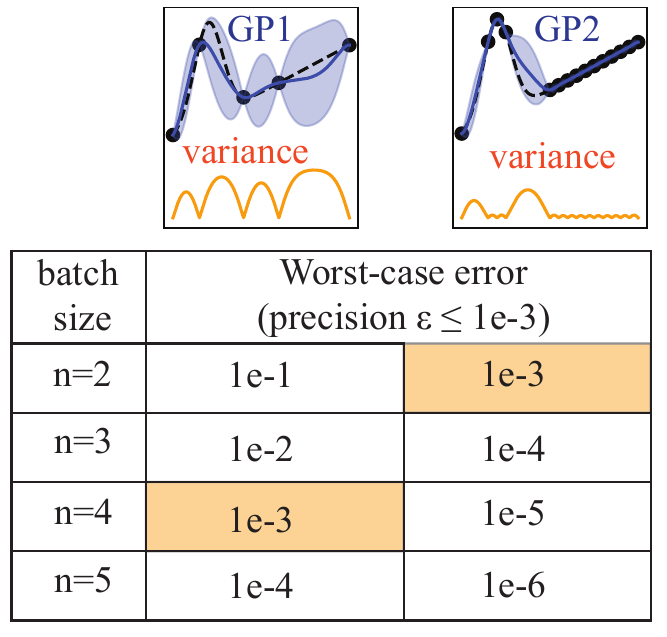}
  \caption{We fix the quadrature precision instead of batch size. The batch size changes adaptively to meet the predefined precision requirement. Our method, AdaBatAL, efficiently determines the optimal number of batch sizes and their querying positions without requiring a brute-force search of all possible batch sizes. AdaBatAL also offers adaptive batch sizes for constrained active learning and constrained Bayesian optimization.}
  \label{fig:concept_batch}
  \vspace{-1em}
\end{figure}
Active Learning (AL) \citep{settles.tr09} is a machine learning concept where the algorithm selects its training data, which enhances accuracy based on fewer labels. Its use is widespread in deep learning models \citep{gal2017deep, ren2021survey, kirsch2019batchbald} and Gaussian processes (GPs) \citep{houlsby2011bayesian, riis2022bayesian}. Bayesian AL intertwines with Probabilistic Numerics (PN) \citep{hennig2015probabilistic, hennig2022probabilistic}, that reinterprets numerical tasks as Bayesian machine learning. This allows uncertainty to interlink with real-world constraints, improving empirical performance, and algorithmic flexibility. In PN, AL enables sample-efficient procedures, with Bayesian optimization (BO) and Bayesian quadrature (BQ) being key instances, applied in fields like drug discovery \citep{gomez2018automatic}, materials \citep{adachi2021high}, and hyperparameter tuning \citep{feurer2015efficient, wu2020practical}.

AL research can be classified into sequential and batch settings.  While the sequential setting selects the next training data point one by one, the batch setting selects multiple points at the same time.
%and assumes the overhead of multiple queries is negligible compared to the expensive labeling process.
We have two key metrics of performance: the number of iterations and the number of total queries. The number of iterations corresponds to the speed of model training, and the batch setting is advantageous as it can gain more feedback per iteration. In contrast, the number of total queries corresponds to the cost. For instance, labeling the data may involve expensive human evaluations. This total query metric is advantageous to the sequential setting as it can observe feedback for every single query to give a rational decision, whereas the batch setting needs to select multiple points without feedback.

However, situations arise where a balance between speed and cost is desirable. For instance, while training a model, renting cloud servers is an option, with charges applied based on the number of nodes (batch size) \textit{and} duration (total queries). Another scenario is crowdsourcing annotation, where a balance is needed between the number of annotators (batch size) and the total working time (total queries). We aim to expedite model training while also saving on cost.

In addition to these situations, constraints often come into play in real-world applications, and often the constraints are also unknown a priori. Unknown constraints \citep{gelbart2014bayesian, hernandez2016a} are the constraints with which we must comply, but we do not know the constraint function a priori and are only observable pointwise. Hence, we had to estimate the true constraint function based on limited observations, resulting in uncertainty in the constraint estimation.
For example, drug discovery needs to satisfy the safety constraints via animal experiments \citep{lipinski1997experimental}, but we do not know the functional form. Similarly, active learning with real physical experiments contains unknown constraints such as limitations from experimental apparatus or phase transition of measuring materials \citep{khatamsaz2023bayesian, lookman2019active}. Training models on the cloud server may be halted due to errors or memory overflow, or annotators may pause annotation in cases of ambiguity in annotation guidelines or unclear samples. Avoiding querying samples that are likely to violate such unknown constraints is essential for the smooth execution of active learning.
However, research on active learning under constraints is scarce, and no existing work considers adaptive batch size under constraints.

To address the said challenges, we propose a PN framework that adaptively adjusts the batch size. Figure \ref{fig:concept_batch} illustrates the concept.
We hypothesize that an adaptive batch size can balance the trade-off between cost and speed. Fixed batch sizes might be ineffective because, as the shape of the acquisition function changes dynamically, the effectiveness of batch acquisition also shifts.
%For instance, multimodal functions require many batch points, while unimodal functions require fewer.
In Figure \ref{fig:concept_batch}, the left side displays four distinct peaks, indicating that four batch sizes would be suitable. Conversely, the right side exhibits only two prominent peaks, suggesting that two batch sizes would be more appropriate.

Given this intuition, we define batch construction as an approximation of a continuous target distribution (e.g., an acquisition function) using a discrete distribution (batch samples)—a process known as \emph{quantization} applied in diverse machine learning fields \citep{graf2007foundations, karvonen2019kernel, teymur2021optimal}. The error in this approximation can be measured by the divergence between the target and the `quantized' distribution. With this perspective, instead of fixing the batch size, we propose to fix the \textit{precision} of the approximation. We reframe batch construction as a quantization task, assessing precision through divergence. We fix the precision requirement for iterations, allowing the batch size and locations to be adaptively adjusted. In essence, our approach quantifies numerical errors stemming from an insufficient batch sizes, strategically harnesses this computational uncertainty for decision-making, and embodies the essence of PN principles.
Specifically, for GP models, this quantization links seamlessly to kernel quadrature (KQ), enabling the use of advanced KQ methods for efficient solutions. As such, we further re-cast the quantization task as a KQ task, using the worst-case integration error as our divergence metric. Our method, \textit{adaptive batch active
learning} (AdaBatAL), efficiently determines the optimal number of batch sizes
and their querying positions without requiring a brute-force search of all possible batch sizes.

AdaBatAL also seamlessly handles AL in the presence of unknown constraints. We view the risk associated with these constraints as a 'varying precision requirement.' If querying points violate the constraints, we remove them from the valid dataset, thereby reducing precision. We interpret a high risk of constraint violation as a lower precision requirement and vice versa. Therefore, the constrained case serves as a 'preprocessing' step to determine the appropriate precision for AdaBatAL, with the constraint model estimating the precision requirement. The versatility of AdaBatAL provides a plug-and-play framework for AL, BQ, and BO, whether constraints are involved or not.

\newpage
\paragraph{Contributions}
%Our contributions are summarised as follows:
\begin{compactenum}
    \item \textbf{Adaptive batch size} We fixed \textit{quadrature precision} via re-casting batch construction as a KQ, allowing batch size adaptively changing according to the acquisition function efficiently.
    \item \textbf{Unknown constraints} We reinterpret the batch AL under unknown constraints as varying precision requirement. This allows adaptively changing the batch size and locations in accordance with the risks of constraint violation.
    \item \textbf{Generality} Our adaptive batch construction scheme applies to AL, BO, and BQ by changing the target distribution of quantization with KQ. Moreover, it applies to non-continuous domains (e.g. combinatorial, mixed feature spaces).
    \item \textbf{Significant improvement} is shown in both batch AL and batch BO tasks, outperforming 17 baselines over 6 synthetic and 7 real-world tasks.
    \item \textbf{Open-source} we open-source the software on GitHub \url{https://github.com/ma921/AdaBatAL}.
\end{compactenum}

\section{Background} \label{sec:background}
We start by providing the background on quantization and KQ. We then demonstrate the connection between GP, KQ, and BQ, leading to pure batch uncertainty sampling. We defer the background of GP and fully Bayesian GP (FBGP) in Supplementary~\ref{sup:gp}.
%\subsection{Kernel quadrature as Quantization}\label{sec:kq}
%We demonstrate the close connection between KQ and the quantization approach widely adopted in batch AL.

\textbf{Quantization.} 
Let $\mu$ be a probability distribution defined on a set $\mathcal{X}$. The \textit{quantization} task is to find the discrete distribution $\nu:= \frac{1}{n} \sum_{i=1}^n \delta_{x_i}$, which best approximates $\mu$ with $n$ representative points $x_i$. Here, $\delta_{x}$ denotes a point mass (delta distribution) located at $x \in \mathcal{X}$. To solve the quantization task, one first identifies an optimality criterion, typically a notion of \textit{discrepancy} between $\mu$ and $\nu$, and then develops an algorithm to approximately minimize it. %When the Wasserstein distance employed, it is known to be the Voronoi partitions \citep{graf2007foundations}.%, which elegantly connects with Voronoi partitions whose centers are $x_i$ \citep{graf2007foundations}.

\textbf{Kernel Quadrature.} 
KQ is a numerical integration for calculating the integral of a function belonging to a reproducing kernel Hilbert space (RKHS).
The aim is to find a good approximation of an, otherwise intractable, integral with a weighted sum. A KQ rule, $Q_{\boldsymbol{w}, \boldsymbol{x}}$, is given by weights $\boldsymbol{w} = \{w_i\}^n_{i=1}$ and points $\boldsymbol{x} = \{ x_i \}^n_{i=1}$,
\begin{equation}
    Q_{\boldsymbol{w}, \boldsymbol{x}}(f) := \sum_{i=1}^n w_i f(x_i) \approx \int_\mathcal{X} f(x) \text{d} \mu(x), \label{eq:kq}
\end{equation}
where $f$ is a function of RKHS $\mathcal{H}$ associated with the kernel $K$. 
%We define its worst-case error by
The \emph{worst-case error} given $\mu$ and $\mathcal{H}$ is
\begin{equation}
    \text{wce}(Q_{\boldsymbol{w}, \boldsymbol{x}}) := \sup_{\lVert f \rVert_{\mathcal{H} \leq 1}} \Big\lvert Q_{\boldsymbol{w}, \boldsymbol{x}}(f) - \int_\mathcal{X} f(x) \text{d} \mu(x) \Big\rvert. \label{eq:wce}
\end{equation}
%Therefore, the goal of KQ is to find the KQ rule $Q_{\boldsymbol{w}, \boldsymbol{x}}$ that minimizes the worst-case error.
The aim is to find $Q_{\boldsymbol{w}, \boldsymbol{x}}$ minimizing worst-case error.

\textbf{Connection to quantization.} %\label{sec:mmd}
When inspecting the KQ rule as integration against a discrete distribution $\nu:= \sum_{i=1}^n w_i \delta_{x_i}$, namely, $Q_{\boldsymbol{w}, \boldsymbol{x}}(f) = \int_\mathcal{X} f(x) \text{d} \nu(x)$, the worst-case error can be viewed as the \textit{divergence} between $\mu$ and $\nu$.
Indeed, there is a theoretical connection between KQ and quantization, as KQ is the \textit{weighted} quantization under the maximum mean discrepancy (MMD) metric \citep{karvonen2019kernel, teymur2021optimal}. MMD is a widely used method to quantify the divergence between two distributions \citep{sriperumbudur2010hilbert, muandet2017kernel}, defined as:
\begin{align*}
    \text{MMD}_\mathcal{H}(\nu, \mu) := \Bigg\lVert \int K(\cdot, x)\text{d}\nu(x) - \int K(\cdot, x)\text{d}\mu(x) \Bigg\rVert_\mathcal{H},
\end{align*}
and we can rewrite as \citep{huszar2012optimally}:
\begin{align*}
    \text{MMD}^2_\mathcal{H}(\nu, \mu) := \sup_{\lVert f \rVert_\mathcal{H} = 1} \Bigg\lvert \int f(x) \text{d} \nu(x) - \int f(x) \text{d} \mu(x) \Bigg\rvert^2.
\end{align*}
This squared formulation is the same with the worst-case error. Therefore, solving KQ is equivalent to finding the discrete distribution $\nu$ that best approximates $\mu$ with regard to MMD. Note that KQ is a weighted quantization, unlike in the previous section.

\textbf{Connection to Gaussian Process.} %\label{sec:gp}
Assume a function $f$ is modelled by GP, $f \sim \mathcal{GP}(m,C)$, with limited number of observed points, $\mathcal{D}_0 := \{\boldsymbol{x}_0, \boldsymbol{y}_0 \}$, where $\boldsymbol{y}_0 = f_\text{true}(\boldsymbol{x}_0) + \epsilon$ are the noisy observations. We wish to estimate the expectation of the function $\hat{Z}:= \int_\mathcal{X} f(x) \text{d} \mu(x)$. This setting is called Bayesian quadrature (BQ) \citep{o1991bayes}, one of the central methods of PN. The integral estimate are as follows:
\begin{align}
    \mathbb{E}_f [ \hat{Z}] 
    &= \int \! m(x)\text{d} \mu(x)
    = \boldsymbol{z}^\top \boldsymbol{K}^{-1} \boldsymbol{y}_0,
\label{eq:bq_mean}\\
    \mathbb{V}\text{ar}_f [ \hat{Z} ]
    &\!=\! \int\! C(x, x^\prime) \text{d} \mu(x) \text{d} \mu(x^\prime)
    = z^\prime - \boldsymbol{z}^\top \boldsymbol{K}^{-1} \boldsymbol{z},
\label{eq:bq_var}
\end{align}
where $\boldsymbol{z} := \int K(x, \boldsymbol{x}_0)\text{d} \mu(x)$ and $z^\prime := \int K(x, x^\prime) \text{d} \mu(x) \text{d} \mu(x^\prime)$ are kernel mean and variance, respectively (see details in Supplementary~\ref{sup:bq})

\citet{huszar2012optimally} proved the worst-case error (Eq.~\labelcref{eq:wce}) equals to the variance in  Eq.~\labelcref{eq:bq_var} if quadrature nodes are $\mathcal{D}_0$. 
%Before following this proof, let us be reminded of the similarities between BQ and KQ.
BQ expectation in Eq.~\labelcref{eq:bq_mean} is a weighted sum; $ \boldsymbol{z}^\top \boldsymbol{K}^{-1} \boldsymbol{y}_0 = \sum_{i=1}^n w_{BQ, i} y_i$, 
where $w_\text{BQ, j} := \sum_{i=1}^n \boldsymbol{z}^\top_{i} \boldsymbol{K}^{-1}_{i,j}$. 
We can further think these weights as a discrete distribition $\nu_\text{BQ} := \sum_{i=1}^n w_{\text{BQ}, i} \delta_{x_i}$, then the variance of integral estimation becomes:
\begin{equation}
    \mathbb{V}\text{ar}_f [ \hat{Z} ]
    = \text{MMD}^2(\mu, \nu_\text{BQ})
    = \inf_{\boldsymbol{w}} \text{wce}(Q_{\boldsymbol{w}, \boldsymbol{x}})^2. \label{eq:kq-eq}
\end{equation}
This shows that KQ and BQ are closely connected (see details in \citep{huszar2012optimally}). The variance of integral is the uncertainty of GP over $\mu$, so quantizing this distribution using KQ can be understood as a `pure batch exploration' of GP uncertainty. This idea was applied to batch BQ \citep{adachi2022fast}.%, which shows both empirically and theoretically strong results.

\textbf{In summary.} A quantization task can be viewed as a KQ task. The selected batch samples minimize the divergence between the target distribution $\mu$ and the batch samples $\nu$, with a given kernel $K$. When we use the GP predictive covariance $C(\cdot, \cdot)$ as the kernel $K$ for the MMD, the KQ becomes the pure batch exploration of GP uncertainty while also minimizing the divergence from the target distribution. Hence, batch construction via solving KQ can offer a quantization of the target distribution combined with uncertainty sampling.

\section{Adaptive Batch Active Learning}\label{sec:abc}
Now, we introduce our method, AdaBatAL. Any KQ method can be used, but we employ the recombination \citep{hayakawa2022positively} for flexibility. We extend this to adaptive batch size under unknown constraints.

\subsection{Problem Setting of Batch Active Learning}
\textbf{Batch Active Learning} 
Consider we have a limited number of a labelled dataset $\mathcal{D}_0 = \{ \boldsymbol{x}_k, \boldsymbol{y}_k\}_{k=1}^m$, and the large number of unlabelled pool set $\mathcal{X}_N = \{ \boldsymbol{x}_l\}_{l=1}^N$, where $N \gg m$, an oracle can provide labels $\mathcal{Y}_N = \{ \boldsymbol{y}_m \}_{m=1}^N$ for the corresponding inputs. We sequentially query the batch samples $\mathcal{D}^n_t  = \{ \boldsymbol{x}_j, \boldsymbol{y}_j\}_{j=1}^n$ with $n$ batch sizes at $t$-th iteration, resulting in the total labelled dataset $\mathcal{D}_t = \mathcal{D}_{t-1} \cup \mathcal{D}^n_t = (\mathcal{X}_t, \mathcal{Y}_t)$, and repeat $T$ times\footnote{To clarify, $D_0 \subseteq D_t$ but $D_0 \not\subset D^n_t$.}.
The batch AL task is to select the $\mathcal{D}_t$ to minimize the prediction error between true labels $\mathcal{Y}_N$ and the prediction conditioned on $\mathcal{X}_N$ at given budget $T$. Throughout this paper, we assume the model is an FBGP for AL and a normal GP for BO.

Following the works \citet{pinsler2019bayesian, adachi2023sober}, we can recast the batch AL and batch BO as a quantization task. The difference between AL and BO comes down to the target distributions $\mu$: the candidate pool of unlabelled inputs for batch AL, and the probability of global optimum location for batch BO. How to recast these tasks to a quantization is not a primary focus of this paper, we defer the explanation of their attempts in Supplementary \ref{sup:acs} and \ref{sup:sober}. Important takeaways from their works are that the quantization approach can outperform popular baselines, such as BALD \citep{houlsby2011bayesian} for batch AL, and hallucination \cite{azimi2010batch} for batch BO. Yet, their approach only considers fixed batch size without constraints. We augment their approaches by adaptive batch sizes under unknown constraints.

\textbf{Unknown Constraints} 
Consider our labelling scheme is subject to the constraint $c(x) \geq 0$, where $c$ is the constraint with which we must comply, otherwise the query $x$ is eliminated from the labelled dataset $\mathcal{D}_t$ (e.g., a drug candidate that breaches a safety constraint will not be tested.). We further assume the constraints are unknown a priori and are only observable pointwise. Hence, probabilistic model estimates the function $\hat{c}(x)$ with its predictive uncertainty, providing the probability of constraint satisfaction $q(x)$ at given input $x$. Following \citet{gelbart2014bayesian}, we model the constraint by another GP (see Supplementary \ref{sup:cBO}).

\subsection{Problem Setting of Kernel Quadrature}
As a general situation, consider we are given a kernel $K$ on $\mathcal{X}$
and an $N$-point samples $\textbf{X}_\text{cand} \in \mathcal{X}^N$ associated with a nonnegative weight $\textbf{w}_\text{cand}$ with $\textbf{w}_\text{cand}^\top\boldsymbol{1} = 1$\footnote{$\boldsymbol{1}$ is $[1,\dotsc,1]^N$, the vector of ones.}. We denote this as $\mu(x) := \sum_{i=1}^N w_i \delta_{x_i}$ as a discrete distribution, or $(\textbf{w}_\text{cand}, \textbf{X}_\text{cand})$ as the ordered pair.
In a typical batch AL setting, $\mu$ is the candidate pool of unlabelled inputs with equal weights. 
%In the batch BO setting, this can be weighted samples depending on the definition of $\mu$.
The goal is to find a weighted subset $(\textbf{w}_\text{batch}, \textbf{X}_\text{batch})$, $\nu(x) := \sum_{j=1}^n w_j \delta_{x_j}$ which minimizes $\text{MMD}_\mathcal{H}(\mu, \nu)$ given $\mu$ and kernel $K$\footnote{We set the kernel $K$ as the posterior predictive covariance $C(\cdot, \cdot)$ (recall the background section).}. Hence, this is a KQ task. The quantized subset $\nu$, $\textbf{X}_\text{batch} \subset \textbf{X}_\text{cand}$, will give the batch samples for batch AL and batch BO.
Unlike the existing setting \citep{hayakawa2022positively,adachi2022fast,adachi2023sober},
we additionally work under the following conditions:
\begin{compactenum}
    \item[(a)] The upper bound of batch size $n$ is given but the actual batch size is adaptively changed to meet the precision under the given tolerance $\epsilon_\text{LP}$.
    \item[(b)] After we choose the batch querying points, $(\textbf{w}_\text{batch}, \textbf{X}_\text{batch})$,
    each point $x\in\textbf{X}_\text{batch}$ is subject to the probabilistic constraint $q(x)$\footnote{The true constraint $c(x)$ is deterministic but $q(x)$ becomes probabilistic due to the predictive uncertainty.} (and violated w.p. $1-q(x)$), where $q:\mathcal{X}\to[0, 1]$ is given as GP. We query the true constraint $c(x)$, then we obtain the feasible points and corresponding weights $(\tilde{\textbf{w}}_\text{batch}, \tilde{\mathbf{X}}_\text{batch})$, where $\tilde{\mathbf{X}}_\text{batch} \subset \mathbf{X}_\text{batch}$\footnote{$\tilde{\mathbf{X}}_\text{batch} = \mathbf{Z}^\top \mathbf{X}_\text{batch}$, where $\mathbf{Z}$ is a vector of Bernoulli random variables with probabilities $q(\mathbf{X}_\text{batch})$}. 
    We use the feasible points for the quadrature.
    \item[(c)] Additionally, a reward function $g:\mathcal{X}\to\mathbb{R}$ is given as additional flexibility that incorporates the other desideratum (e.g. soft constraint), and we want to make the expected reward $\tilde{\textbf{w}}_\text{batch}^\top g(\tilde{\textbf{X}}_\text{batch})$
    as big as possible
    while making the worst-case error $\text{wce}(Q_{\tilde{\textbf{w}}_\text{batch}, \tilde{\textbf{X}}_\text{batch}})$ \footnote{For brevity, b is batch, c is cand, then 
    $\text{wce}(Q_{\tilde{\textbf{w}}_\text{b}, \tilde{\textbf{X}}_\text{b}}) 
    = \tilde{\textbf{w}}_\text{b}^\top K(\tilde{\textbf{X}}_\text{b}, \tilde{\textbf{X}}_\text{b}) \tilde{\textbf{w}}_\text{b} - 2 \tilde{\textbf{w}}_\text{b}^\top K(\tilde{\textbf{X}}_\text{b}, \textbf{X}_\text{c}) \textbf{w}_\text{c} + \textbf{w}_\text{c}^\top K(\textbf{X}_\text{c}, \textbf{X}_\text{c}) \textbf{w}_\text{c}$.
    } as small as possible.
\end{compactenum}

\subsection{Kernel Quadrature via Nyström Approximation}\label{sec:nystrom}
Although the Nyström method \citep{williams2000using, drineas2005on, kumar2012sampling} is primarily used for approximating a large Gram matrix by a low-rank matrix, it can also be used for directly approximating the kernel function itself.
Given a set of $M$ points $X_\text{nys} = \{ x_i \}_{i=1}^M \subset \mathcal{X}$, the Nyström approximation of $K(x, y)$ is given by:
\begin{equation}
    K(x, y) \approx K_0(x, y) := \sum_{i=1}^{n-1} \lambda_i^{-1}\varphi_i(x)\varphi_i(y),\label{eq:nys}
\end{equation}
where $\varphi_i(\cdot) := u^\top_i K(X_\text{nys}, \cdot)$ $(i=1,\dotsc,n-1)$ are called \emph{test functions}, chosen from a larger $M$ dimensional space $\text{span}\{ K(x_i, \cdot)\}_{i=1}^M$.
The Eq.~\labelcref{eq:nys} holds if $\lambda_s > 0$.
To compute Eq.~\labelcref{eq:nys}, we perform the best rank-$s$ approximation of the Gram matrix $K(X_\text{nys}, X_\text{nys})= U \Lambda U^\top$, given by eigendecomposition, where $U = [u_1,\dotsc,u_M] \in \mathbb{R}^{M \times M}$ is a real orthogonal matrix and $\Lambda = \text{diag}(\lambda_1, \dotsc ,\lambda_M)$ with $\lambda_1 \geq \dotsc \geq \lambda_M \geq 0$. 

We can use the test functions for integration estimator $\hat{Z} = \int_\mathcal{X} f(x) \text{d} \mu(x)$. When the spectral decay in eigenvalues is steep, the Nyström method can give a good approximation of the original kernel function with a small number of test functions.
Let $\boldsymbol\varphi = \{\varphi_1,\dotsc,\varphi_{n-1} \}^\top$ be the vector of test functions that spans $\mathcal{H}_{K_0}$, the RKHS associated with the approximated kernel $K_0$, we assume we have additional knowledge of expectations, namely, $\int_\mathcal{X} \boldsymbol\varphi(x) \text{d} \mu(x) = \textbf{w}_\text{cand}^\top \boldsymbol\varphi(\textbf{X}_\text{cand})$ is given.  
%and let $\mu(x)$ be the probability measure we wish to integrate.
We can actually construct a convex quadrature $Q_n = (w_i, x_i)^n_{i=1}$:
\begin{align}
\sum_{i=1}^{n-1} w_i \varphi_i(x_i) = \int_\mathcal{X} \boldsymbol\varphi(x) \text{d} \mu(x) \approx \int_\mathcal{X} f(x) \text{d} \mu(x). \label{eq:testf}
\end{align}
Now, we can approximate the integral by $n-1$ test functions.
Hence, Eq.~\labelcref{eq:testf} can be understood as $n-1$ \emph{equality constraints} which $w_i$ and $x_i$ need to satisfy.

The benefit of this approximation is to incorporate the information of spectral decay of Gram matrix for faster convergence. If the target function $f$ is smooth, the spectral decay is fast, then the small number of test functions can well represent the function, leading to batch-size efficient AL and BO \citep{hayakawa2022positively, adachi2022fast}.

\subsection{Linear Programming Formulation}\label{sec:LP}
To solve the above problem, we introduce the following linear programming (LP) problem that aims to achieve both the reward maximization and the worst-case error minimization where possible,
given by modifying the algorithm adopted in \citep{adachi2023sober} ($n\ge 3$):
\begin{align}
    \begin{array}{rl}
    &\underset{\textbf{w}}{\text{maximize}} \quad
    \textbf{w}^\top \bigl[ g(\textbf{X}_\text{cand}) \odot q(\textbf{X}_\text{cand})\bigr], \\
    &\underset{\phantom{a}}{\text{subject to}}\\
    & \bigl\lvert(\textbf{w} - \textbf{w}_\text{cand})^\top \varphi_j(\textbf{X}_\text{cand}) \bigr\rvert \leq \epsilon_\text{LP}\sqrt{\lambda_j / (n - 2) },\nonumber\\
    &\forall j: 1 \le j \le n - 2 ,\\
    &(\textbf{w} - \textbf{w}_\text{cand})^\top q(\textbf{X}_\text{cand}) \ge 0,\\ 
    &\textbf{w}^\top \textbf{1} = 1, \ \ 
    \textbf{w} \geq \textbf{0},\ \ 
    |\textbf{w}|_0 \le n,
    \end{array}
    \label{eq:lp}
\end{align}
where $\epsilon_\text{LP}\ge 0$ is a {\it tolerance} parameter, which can be interpreted as the quadrature precision requirements (smaller is more accurate), and
$(\lambda_j, \varphi_j)$ are given by the Nyström approximation (see \ref{sec:nystrom})\footnote{$\odot$ refers to Hadamard product, and $|\cdot|_0$ denotes the number of nonzero entries.}.

The intuition of this formulation is as follows:
\begin{compactenum}
    \item[(1)] The solutions are the sparse weights $\textbf{w}$, where the non-zero element of $\textbf{w}$ corresponds to the batch selection, and the corresponding samples of $\textbf{X}_\text{cand}$ is the batch samples $\textbf{X}_\text{batch}$. We refer to the nonzero weights and corresponding samples as the solution $(\textbf{w}_\text{batch}, \textbf{X}_\text{batch})$\footnote{$\textbf{X}_\text{batch} \subset \textbf{X}_\text{cand}$, $\textbf{w}_\text{batch} \subset \textbf{w}_\text{cand}$, and $|\textbf{X}_\text{batch}| = |\textbf{w}|_0$}, and its batch size is  $|\textbf{X}_\text{batch}| \leq n$. As such, this LP problem is to subsample the batch samples $\nu$ from the given discrete distribution $\mu$, namely, quantization.
    \item[(2)] The objective is to maximize the expected reward $g$ under the risk of constraint violation $q$. This promotes safe sampling by increasing the expected constraints satisfaction $\textbf{w}^\top q(\textbf{X}_\text{cand})$.
    \item[(3)] The first constraints correspond to equality constraints with test functions in Eq.~\labelcref{eq:testf}. We relaxed the equality constraints to inequality constraints to accept the tolerance $\epsilon_\text{LP}$. These $n-2$ inequality constraints restrict the solution space to where the approximation error of the expectations of test functions $\lvert (\textbf{w} - \textbf{w}_\text{cand})^\top  \varphi_j(\textbf{X}_\text{cand}) \rvert$\footnote{This is a quadrature error in Eq.~\labelcref{eq:testf}. $\bigl\lvert \textbf{w}^\top \boldsymbol\varphi(\textbf{X}_\text{cand}) - \textbf{w}_\text{cand}^\top \boldsymbol\varphi(\textbf{X}_\text{cand}) \bigr\rvert \approx \bigl\lvert \int_\mathcal{X} f(x) \text{d} \nu(x) - \int_\mathcal{X} f(x) \text{d} \mu(x) \bigr\rvert$.} is within the tolerance parameter $\epsilon_\text{LP}$. These $n-2$ constraints are very restrictive; the flexibility to select the larger objective is much more restricted than the typical LP problem. $\epsilon_\text{LP}$ controls the trade-off between the accuracy for quadrature and relaxing solution space to find the larger objective.
    \item[(4)] Other constraints assure the number of nonzero elements of the solution set $\textbf{w}$ is fewer than the upper bound of batch size $n$, the convex and positive weights, and the probability of probabilistic constraints' satisfaction is positive.
\end{compactenum}
Thus, in response to conditions (b)(c), the solution of this LP problem provides the batch samples that satisfy convex quadrature rules within the tolerance \textit{and} maximizing the reward. The balance between quadrature accuracy and reward maximization can be controlled by a single parameter $\epsilon_\text{LP}$. To be clear, only within §\ref{sec:LP}, the term `constraints' refers to the ones in LP formulations. Otherwise, the constraints refer to the task-specific unknown constraints (e.g. safety constraints for drug discovery).

\subsection{Adaptive Batch Sizes}\label{sec:adaptive_batch_sizes}
The count of non-zero elements, denoted as $|\boldsymbol{w}|_0$, is adjusted based on the tolerance $\epsilon_\text{LP}$. The intuition of the batch size adaptivity is explicated as:
\begin{compactenum}
    \item Higher precision demands result in a smaller quadrature error tolerance. This necessitates a larger sample set for more precise integration.
    \item Conversely, lower precision requirements needs fewer     $|\boldsymbol{w}|_0$ to meet the desired accuracy.
\end{compactenum}
Elaborating further, the batch size is tied to slack variables in LP solvers. As the tolerance $\epsilon_\text{LP}$ increases, some inequality constraints become deactivated \citep{dantzig2002linear}. The batch size is determined by the number of active constraints, often leading to sparse weights with $|\boldsymbol{w}|_0 < n$. When constraints are loose, a large preset batch size is inefficient, as the desired precision can be achieved with fewer samples. As such, we can identify the adaptive batch size $|\boldsymbol{w}|_0$ without needing a brute-force search of all possible batch sizes.

Note that $\epsilon_\text{LP}$ controls \emph{all} balances: the batch size, quadrature accuracy, and reward maximization. Interestingly, its behavior is not a monotonic decrease in its magnitude. As $\epsilon_\text{LP}$ approaches infinity, the batch size converges to 1, aligning with the sequential AL case. An increased $\epsilon_\text{LP}$ shrinks the batch size as observed in §\ref{sec:adabatch}. This approach is essentially a heuristic for adaptive batch sizes. Although it satisfies a predefined worst-case error threshold, it does not guarantee optimal results based on other established metrics like mutual information \citep{krause2012near}. However, as \citet{leskovec2007cost} highlighted, when greedily maximizing mutual information under the weighted candidates and a budget constraint (limitation in the number of the total queries), the approximation factor can be arbitrarily bad. Hence, even popular strategies, such as BALD \citep{houlsby2011bayesian}, also cannot achieve a solution within $1 - 1 / e$ of the optimal in our problem setting \citep{li2022batch}.

\subsection{Unknown Constraints As The Lowered Precision Requirement}\label{sec:unknwon_constraints}
\begin{figure}
  \centering
  \includegraphics[width=\hsize]{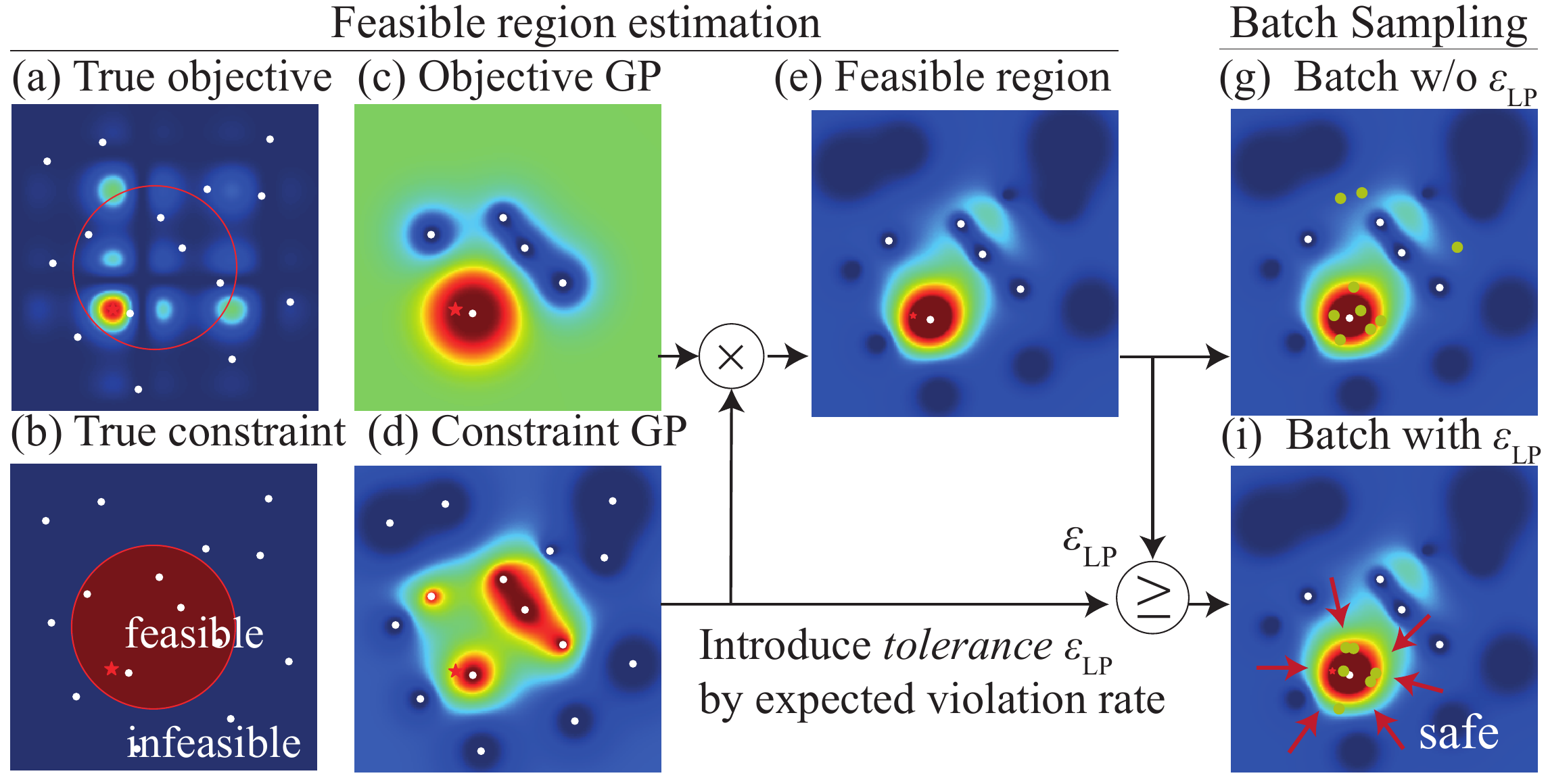}
  \caption{Constrained batch active learning. As the increased violation risk $\epsilon_\text{vio}$ propagates to the tolerance $\epsilon_\text{LP}$, reward maximization is subsequently prioritized over quadrature, resulting in safe batch samples.}
  \label{fig:concept}
  \vspace{-1em}
\end{figure}
In this further examination, we address the probabilistic constraint denoted as $q$. Given the uncertainty in predicting the true constraint $c$, the candidate solution, $\textbf{X}_\text{cand}$, carries a risk of violation. We can estimate the \emph{expected} violation rate by $\epsilon_\text{vio}:= 1 - \textbf{w}_\text{cand}^\top q(\textbf{X}_\text{cand})$. It is assumed that infeasible points are eliminated from quadrature nodes for computation, reducing quadrature accuracy. The expected violation rate $\epsilon_\text{vio}$ can be interpreted as the \emph{risk} we cannot control. A high-risk scenario necessitates cautious exploration to avoid wasting valuable queries; this suggests smaller batch sizes and selecting queries where $\textbf{x}_\text{cand}$ is more likely to satisfy the true constraint $c$. Conversely, a low risk allows for more optimistic exploration.

In response to varying risk levels, we advocate for an \emph{adaptive} exploration strategy. Our proposed method is straightforward yet effective: setting $\epsilon_\text{LP} = \epsilon_\text{vio}$. This approach allows for automatic adjustment of exploration safety. When $\epsilon_\text{vio}$ is high, indicating greater risk, $\epsilon_\text{LP}$ is set higher. This results in looser quadrature precision, smaller batch sizes, and a solution space that is more likely to satisfy constraints\footnote{Remember that a reduction in quadrature precision results in an expansion of the solution space, which in turn enables the identification of solutions with higher LP objective values, as denoted by $\textbf{w}_\text{cand}^\top \bigl[ g(\textbf{X}_\text{cand}) \odot q(\textbf{X}_\text{cand})\bigr]$, where $\textbf{w}_\text{cand}^\top q(\textbf{X}_\text{cand})$ represents the expected satisfaction of the constraint. Thus, maximizing the LP objective value leads to increasing the constraint satisfaction probability.}. Thus, a higher $\epsilon\text{LP}$ leads to safer batch sampling. When the risk $\epsilon_\text{vio}$ is low, $\epsilon_\text{LP}$ is set lower, allowing for larger batch sizes and more explorative solutions. Figure~\ref{fig:concept} demonstrates this adaptive behavior: high-risk information $\epsilon_\text{vio}$ influences $\epsilon_\text{LP}$, leading to safer batch samples.
Adaptive safe exploration is not necessarily always safe. We need to explore uncertain regions at some point and we propose it is when the risk is low. Our PN framework effectively bridges computational uncertainty and real-world risk, providing an automated and adaptable balance between safety and exploration.

\subsection{Error Bounds}
The error estimate of KQ is essentially determined by the approximation error of the Nyström method, 
$\epsilon_\text{nys}:=\max_{x\in\textbf{X}_\text{cand}}
\lvert K_0(x, x) - K(x, x)\rvert^{1/2}$.
Error bounds for this approximation have been well studied in the literature \citep{drineas2005on,kumar2012sampling,hayakawa2023sampling}.

\begin{prop}\label{prop:lp}
    Under the above setting, let 
    \rm$\textbf{w}_*$\it\ be the optimal solution of the LP, %~Eq.~\labelcref{eq:lp},
    and let \rm$\textbf{X}_\text{batch}$\it\ be the subset of \rm$\textbf{X}_\text{cand}$\it, 
    corresponding to the nonzero entries of \rm$\textbf{w}_*$\it\ 
    (denoted by \rm$\textbf{w}_\text{batch}$\it).
    Suppose that \rm$\tilde{\textbf{X}}_\text{batch}$\it\ 
    is given by a random subset of \rm$\textbf{X}_\text{batch}$\it, where each point
    $x$ satisfies the constraints with probability $q(x)$,
    and let \rm$\tilde{\textbf{w}}_\text{batch}$\it\ be the corresponding weights.
    Then, we have
    \rm\begin{equation}
        \mathbb{E}[\tilde{\textbf{w}}_\text{batch}^\top g(\tilde{\textbf{X}}_\text{batch})]
        \ge \textbf{w}_\text{cand}^\top \bigl[ g(\textbf{X}_\text{cand}) \odot q(\textbf{X}_\text{cand})\bigr],
        \label{eq:lp-1}
    \end{equation}\it
    and, for any function $f$ in the RKHS with kernel $K$,
    %we have
    \rm\begin{align}
    \begin{split}
        &\mathbb{E}\!\left[\left\lvert \tilde{\textbf{w}}_\text{batch}^\top
        f(\tilde{\textbf{X}}_\text{batch}) -
        \textbf{w}_\text{cand}^\top f(\textbf{X}_\text{cand})\right\rvert\right]\\
        &\le
        (\epsilon_\text{vio} K_{\max}
        + 2\epsilon_\text{nys} + \epsilon_\text{LP})\lVert f \rVert,
        \label{eq:lp-2}
    \end{split}
    \end{align}
    \it
    where $\lVert f \rVert$ is the RKHS norm of $f$,
    $K_{\max}:=\max_{x\in\textbf{X}_\text{cand}}K(x, x)^{1/2}$,
    and \rm$\epsilon_\text{vio}:= 1 - \textbf{w}_\text{cand}^\top
    q(\textbf{x}_\text{cand})$\it\ is the expected violation rate
    with respect to the empirical measure given by
    \rm$(\textbf{w}_\text{cand}, \textbf{X}_\text{cand})$\it.
\end{prop}

The proof is given in Supplementary \ref{sec:proof}. This proposition indicates that we can obtain a quantitative estimate of the two tasks described in (c) concurrently. We can attain at least the expected reward of the original batch while ensuring that the resulting measure (which may not necessarily be probabilistic) integrates the functions in the RKHS within a proven error.

\subsection{How to Solve The LP Problem}\label{sec:complexity}
We used Gurobi \citep{gurobi} to solve the LP problem. We used the randomized singular value decomposition to eigendecompose the Gram matrix \citep{hal11} with $M$-point samples $\textbf{X}_\text{nys}\subset\textbf{X}_\text{cand}$. We set $\epsilon_\text{LP} = 10^{-8}$ as the lower bound to avoid LP failure due to the randomness of $\mu$. The complexity of this computation is lower than $\mathcal{O}(N M + M^2 \log n + M n^2 \log (N / n))$ \citep{hayakawa2022positively}.

\textbf{Probability function $q$ } A probability function $q$ can be a given constraint function \citep{gardner2014bayesian}, or estimated function as another GP \citep{gelbart2014bayesian} (see details in Supplementary~\ref{sup:cBO}). If there is no constraints, we can simply set $q(x) = 1$, then it becomes standard batch AL, BQ, or BO.

\textbf{Reward function $g$ } A reward function $g$ is for an additional flexibility to incorporate the information. If we do not have particularly informative information to add, we can simply set as $g=1$. We can view $g$ as the soft constraint of the objective. We can set $g$ for another acquisition function, or prior knowledge of global optimum such as \citet{hvarfner2022pi, adachi2024looping}.

%\textbf{Hyperparameters and Continuous Domain } AdaBatAL is 
%Furthermore, to speed up the algorithm, we can make use of the recursive recombination algorithm \citep{lit12,tch15,maalouf2019fast} instead of rigorously maximising the objective function.
%The extension of the algorithm is essentially the same as in \texttt{SOBER}, but we use an LP solver instead of SVD-based elimination at each recursive step for the introduction of tolerance and better optimisation of the objective function.

\section{Related Work}\label{sec:related}
\paragraph{Batch Active Learning and Optimization}
There are a wide variety of batch methods has been proposed: (1) batch AL; for kernels \citep{kremer2014active, joshi2009multi, leskovec2007cost, riis2022bayesian}, deep learning \citep{gal2017deep, kirsch2019batchbald, sener2017active, pinsler2019bayesian}. (2) batch BQ \citep{wagstaff2018batch, adachi2022fast, adachi2023bayesian}, (3) batch BO, a greedy extension of sequential algorithms \citep{azimi2010batch, gonzalez2016batch, eriksson2019scalable, balandat2020botorch}, diversified batch with determinantal point process (DPP) \citep{kathuria2016batched, nava2022diversified}. Constrained batch sampling has been researched in BO \citep{hernandez2016a, benjamin2019constraned, eriksson2021scalable}. However, most do not discuss the quality of batch construction, like KQ methods. The adaptive batch size setting only found in BO \citep{nguyen2016budgeted}, to the best of our knowledge.

\paragraph{Kernel Quadrature}
There are a number of KQ algorithms; herding/optimization \citep{chen2010super, bach2012on, huszar2012optimally}, random sampling \citep{bach2017on, belhadji2019kernel}, DPP \citep{belhadji2019kernel, belhadji2021analysis}, kernel thinning \citep{dwivedi2021kernel, dwivedi2022generalized}, recombination \citep{hayakawa2022positively, hayakawa2023sampling}, kernel Stein discrepancy \citep{chen2018stein, chen2019stein, teymur2021optimal}, randomly pivoted Cholesky \citep{epperly2023kernel}. While any KQ algorithms can be used to solve our problems, we focused on the recombination algorithm due to its flexibility.

\section{Experiments}\label{sec:experiments}
We evaluate our new algorithm, \emph{AdaBatAL}, on synthetic and real-world tasks on batch AL and BO, with and without probabilistic constraints. We implemented AdaBatAL using PyTorch \citep{paszke2019pytorch}, GPyTorch \citep{gardner2018gpytorch}, BoTorch \citep{balandat2020botorch}, and SOBER \citep{adachi2023sober}. All experiments were averaged over 10 repeats, and performed on a laptop\footnote{Performed on MacBook Pro 2019, 2.4 GHz 8-Core Intel Core i9, 64 GB 2667 MHz DDR4}. We fix the number of initial random samples for objective queries to $n_\text{obj} = 10$. The details on experimental conditions and background on real-world examples are summarized in Supplementary \ref{sup:exp}.

\subsection{Efficacy of Adaptive Batch Size}\label{sec:adabatch}
\begin{figure}
  \centering
  \includegraphics[width=0.9\hsize]{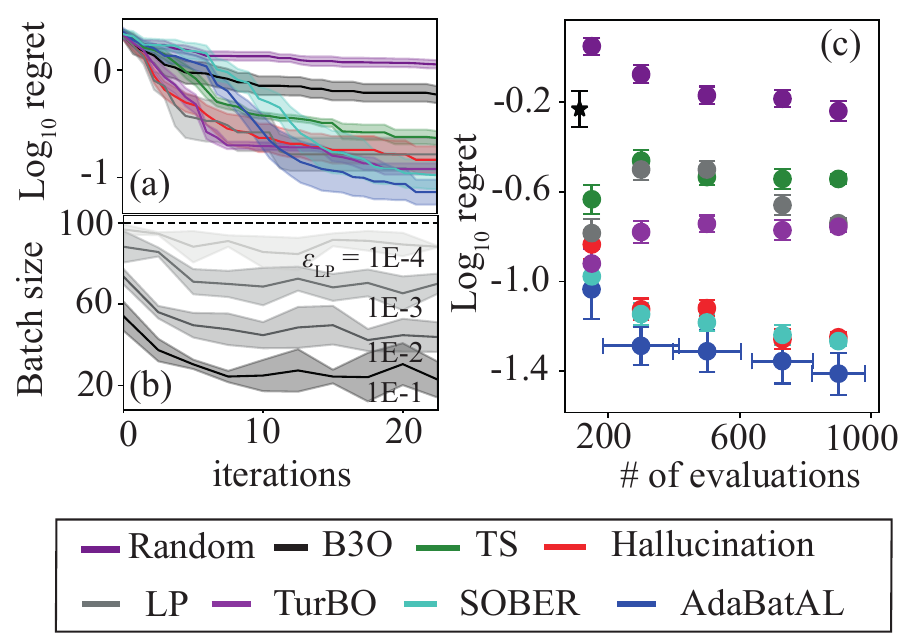}
  \caption{Batch Bayesian optimization results on Hartmann ($d=6$): (a) convergence plot with ($n \leq 5$). (b) batch size variability ($n \leq 100$). The tolerance is set ($\epsilon = 10^{-1}, 10^{-2}, 10^{-3}, 10^{-4}$). (c) Total queries vs. simple regret at the last iteration results of (a)(b). For fixed batch size methods, the mean batch size of AdaBatAL is used ($n = 5, 30, 50, 73, 90$). The plot shows mean $\pm$ standard error of the mean.}
  \label{fig:batchsize}
  \vspace{-1em}
\end{figure}
We first investigate the effect of the adaptive batch size itself \emph{without} unknown constraints, namely, $q(x) = 1$. To compare with the only baseline of the adaptive batch size method, B3O \citep{nguyen2016budgeted}, we selected the batch BO setting. We compared AdaBatAL with the 6 popular baselines of batch BO; B3O, Thompson sampling (TS) \citep{kandasamy2018parallelised}, hallucination \citep{azimi2010batch}, local penalization (LP)\footnote{Only within this section~\ref{sec:adabatch}, LP refers to local penalization. Otherwise, LP means linear programming.} \citep{gonzalez2016batch}, TurBO \citep{eriksson2019scalable}, SOBER \citep{adachi2023sober}. 

Figure~\ref{fig:batchsize} illustrates that AdaBatAL consistently outperformed the baselines throughout the experiments. An increase in the tolerance $\epsilon_\text{LP}$ results in a reduced batch size. Over iterations, the batch size decreases for all values of $\epsilon_\text{LP}$. This indicates that AdaBatAL initially needs more exploratory samples, then it squeezes its search space for exploitation. When matched against fixed batch size methods with a total cost, AdaBatAL achieves a lower regret for the same budget, even when compared to the original SOBER. While B3O tends to opt for a small batch size of around 4, AdaBatAL can adjust its batch size by $\epsilon_\text{LP}$.

\subsection{Efficacy of Expected Violation Rate}
\begin{figure}
  \centering
  \includegraphics[width=0.9\hsize]{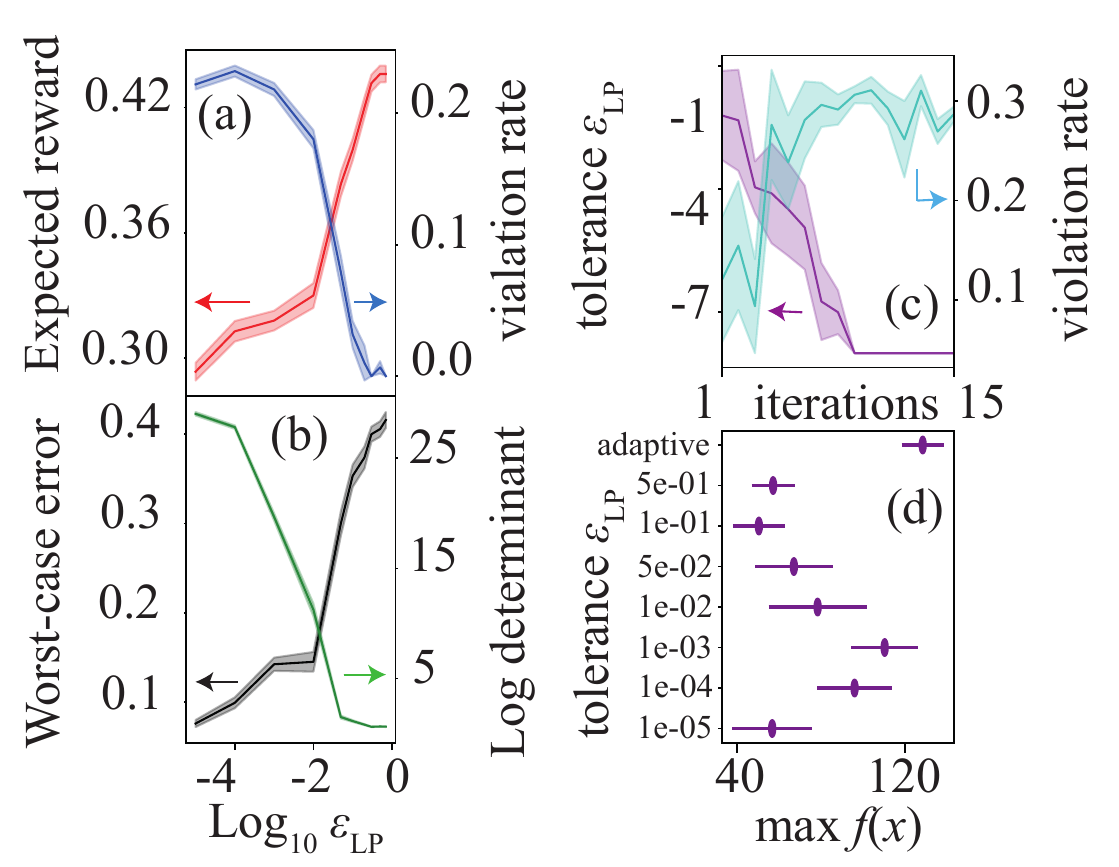}
  \caption{Tolerance effect on constrained batch BO on Branin ($d=2$): the balance between (a) violation rate and expected reward, and (b) worst-case error and log determinant. (c) Tolerance adaptively controls violation rate, and (d) outperforms the fixed cases. (a)(b)(c) are the two Y-axis plots where the color and arrow indicate which Y axis to see.}
  \label{fig:tolerance}
  \vspace{-1em}
\end{figure}
\begin{figure*}[ht!]
  \centering
  \includegraphics[width=0.9\hsize]{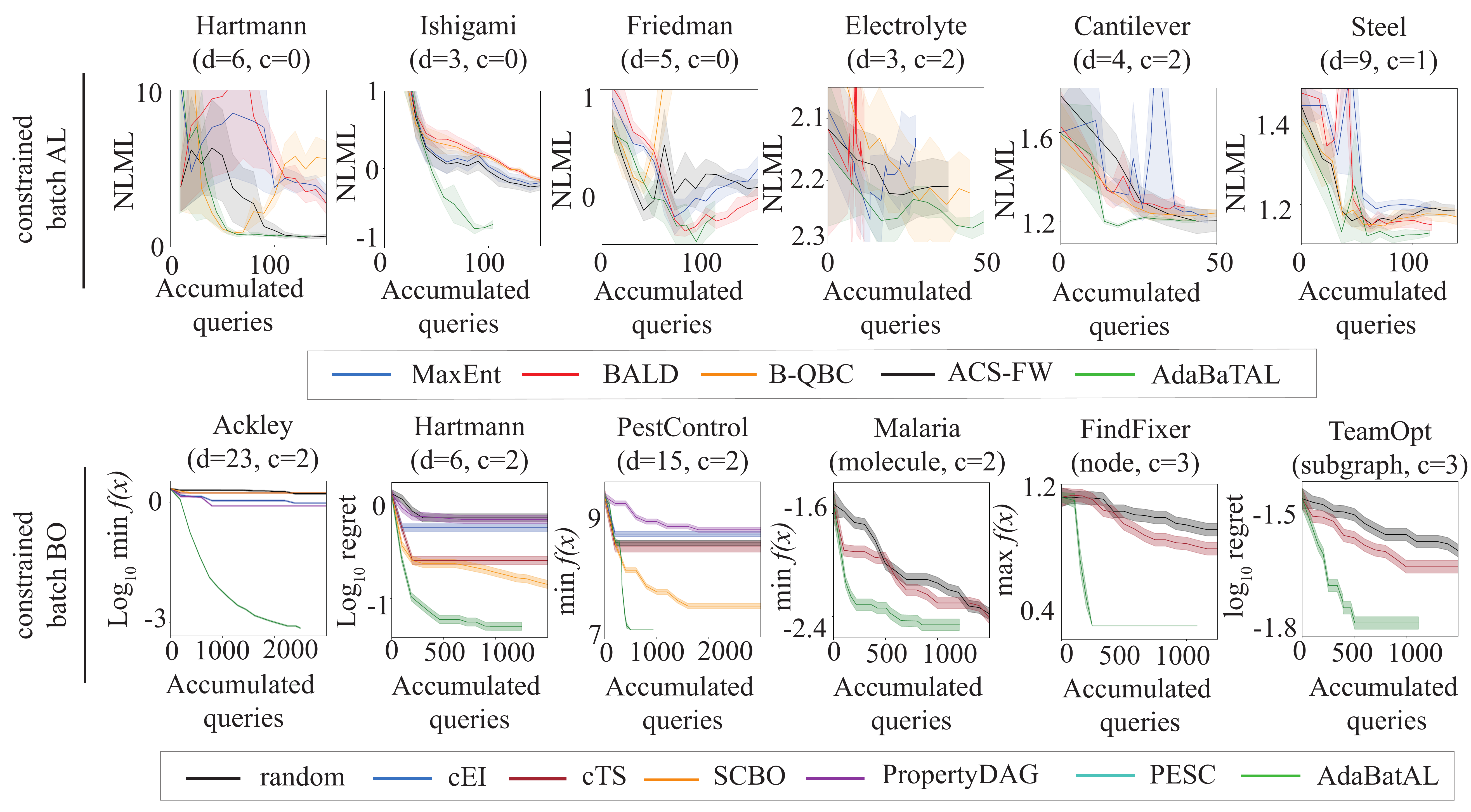}
  \caption{Convergence plot of both constrained batch active learning and Bayesian optimization results across 5 synthetic functions and 7 real-world tasks . $d$ is the dimension, $c$ is the number of unknown constraints. Negative log marginal likelihood (NLML) for active learning tasks, log regret or log best observations for optimization task. Lines and shaded area denote mean ± 1 standard error.}
  \label{fig:batchBO}
  \vspace{-1em}
\end{figure*}
We empirically examine the role of expected violation rate $\epsilon_\text{vio}$ in constrained BO as the time-varying tolerance $\epsilon_\text{LP}$.
Figure \ref{fig:tolerance} presents the main findings.

\textbf{Four key metrics}
\begin{compactenum}
    \item[(1)] \emph{The expected reward} (LP objective): the proxy for how safely we explore.
    \item[(2)] \emph{The violation rate} $1 - \lvert \tilde{\textbf{X}}_\text{batch} \rvert / \lvert \textbf{X}_\text{batch} \rvert$: the proxy for actual results on how safely we explore.
    \item[(3)] \emph{The worst-case error} $\text{wce}(Q_{\tilde{\textbf{w}}_\text{batch}, \tilde{\textbf{X}}_\text{batch}})$: the precision of quadrature.
    \item[(4)] \emph{log determinant} $\log \lvert  K(\tilde{\textbf{X}}_\text{batch}, \tilde{\textbf{X}}_\text{batch}) \rvert$: the proxy for how diversely we explore.
\end{compactenum}

We examined the impact of \(\epsilon_{\text{LP}}\) on four key metrics, as discussed in § \ref{sec:unknwon_constraints}. We aligned $\epsilon_{\text{LP}}$ with $\epsilon_{\text{vio}}$ to facilitate \emph{adaptive} exploration relative to the specified risk level $\epsilon_{\text{vio}}$. The x-axis in Figures (a) and (b) represents variations in $\epsilon_{\text{vio}}$. At higher risk levels, it is essential to prioritize safety. Consequently, there is an increase in the expected reward, correlating with a higher likelihood of constraint satisfaction. This relationship is evident through a reduction in the violation rate, which signifies safer exploration practices.
The numerical metrics give insights of safety exploration in the numerical level: 
High tisk leads to an increase in the worst-case error, which reflects a relaxation in precision requirements. A smaller log determinant suggests less diversity in batch samples, indicated by the proximity of the selected points \(\textbf{X}_{\text{cand}}\) to each other. Conversely, at lower risk levels, we observe a trend towards more optimistic and exploratory sampling.
Hence, our findings confirm that by setting $\epsilon_{\text{LP}} = \epsilon_{\text{vio}}$, our batch exploration successfully adapts to varying risk levels.

We further examined the evolution of the expected violation rate $\epsilon_\text{vio}$ during the optimization loop. As depicted in Figure~\ref{fig:tolerance} (c), the expected violation rate $\epsilon_\text{vio} = \epsilon_\text{LP}$ begins high and diminishes to a minimal value over time. This trend suggests an initial emphasis on safely gathering data, transitioning to greater exploration later on. This approach mirrors strategies like `safe' BO \citep{sui2015safe}, which has demonstrated strong empirical performance (e.g., Figure 4 in \citet{xu2023constrained}) backed by theoretical guarantee. The adaptive tolerance inherently exhibits this behavior with adaptive batch size. Moreover, Figure~\ref{fig:tolerance} (d) indicates that adaptive tolerance converges more rapidly than fixed versions. Notably, the most effective fixed tolerance was $\epsilon_\text{LP} = 10^{-3}$, suggesting that even in the absence of adaptive tolerance, AdaBatAL outperforms the original SOBER ($\epsilon_\text{LP} = 0$) under constraints.

\subsection{Empirical Evaluation}
We tested AdaBatAL's empirical performance across diverse tasks. For batch AL, we compared against five baselines: MaxEnt \citep{mackay1992information}, BALD \citep{houlsby2011bayesian, kirsch2019batchbald}, B-QBC \citep{riis2022bayesian}, and ACS-FW \citep{pinsler2019bayesian}. We evaluated on three synthetic and three real-world tasks. For batch BO, we also explored constrained batch BO and compared against five popular baselines: random, cEI \citep{benjamin2019constraned}, PESC \citep{hernandez2016a}, SCBO \citep{eriksson2021scalable}, and cTS \citep{eriksson2021scalable}. The Malaria, FindFixer, and TeamOpt tasks involve non-continuous inputs over non-Euclidean spaces, each requiring specialized kernels (Tanimoto kernel \citep{ralaivola2005graph} for molecules and the diffusion graph kernel \citep{zhi2023gaussian} for graphs). Due to this unique and crucial real-world setting, the only comparable baselines were random and cTS. Others utilized a standard GP with an RBF kernel. It is important to note that this is \emph{constrained} batch BO, which differs from normal batch BO. Typically, constrained batch BO extends the standard acquisition function with regular batch methods. We chose cEI and cTS as representative methods for these approaches. More details are available in Supplementary \ref{sup:exp}. Figure~\ref{fig:batchBO} shows AdaBatAL's strong empirical performance.

\section{Discussion and Limitations}
We introduced AdaBatAL, a versatile approach capable of adaptive batch sizes under probabilistic constraints for both AL and BO. It is also applicable for non-continuous inputs (e.g., strings for drug discovery and graphs for social data) and arbitrary acquisition functions as the reward function.
%While its empirical performance is promising, more theoretical analysis is desired for a better understanding and potential improvements. For instance, the superiority of subsample-based KQ in practice, compared to its theoretical expectations, remains elusive. Random convex hulls and hypercontractivity \citep{hayakawa2023estimating, hayakawa2022hypercontractivity} offer explanations, albeit in slightly different contexts. 
AdaBatAL is best suited for batch sizes larger than three and does not support asynchronous batch settings \citep{kandasamy2018parallelised}. Its efficacy in high-dimensional BO, which often faces challenges with slow eigenvalue decay, remains an open problem. However, the error bounds of the Nyström method are not directly related to dimensionality; rapid convergence is possible if the function exhibits fast eigenvalue decay, as in the case of the Ackley function.\looseness=-1

\subsubsection*{Acknowledgements}
We thank Samuel Daulton, Siu Lun Chau, Wenjie Xu, and Pierre Osselin for helpful feedback on our paper, and anonymous reviewers who gave useful comments. Masaki Adachi was supported by the Clarendon Fund, the Oxford Kobe Scholarship, the Watanabe Foundation, and Toyota Motor Corporation. Satoshi Hayakawa was supported by the Clarendon Fund, the Oxford Kobe Scholarship, and the Toyota Riken Overseas Scholarship. Harald Oberhauser was supported by the DataSig Program [EP/S026347/1], the Hong Kong Innovation and Technology Commission (InnoHK Project CIMDA), and the Oxford-Man Institute. Martin Jørgensen was partly supported by the Research Council of Finland (grant 356498).

\bibliography{aistats2024_conference}
\bibliographystyle{aistats2024_conference}

 \begin{enumerate}
 \item For all models and algorithms presented, check if you include:
 \begin{enumerate}
   \item A clear description of the mathematical setting, assumptions, algorithm, and/or model. [Yes]
   \item An analysis of the properties and complexity (time, space, sample size) of any algorithm. [Yes in Supplementary]
   \item (Optional) Anonymized source code, with specification of all dependencies, including external libraries. [Not Applicable]
 \end{enumerate}

 \item For any theoretical claim, check if you include:
 \begin{enumerate}
   \item Statements of the full set of assumptions of all theoretical results. [Yes]
   \item Complete proofs of all theoretical results. [Yes]
   \item Clear explanations of any assumptions. [Yes]     
 \end{enumerate}

 \item For all figures and tables that present empirical results, check if you include:
 \begin{enumerate}
   \item The code, data, and instructions needed to reproduce the main experimental results (either in the supplemental material or as a URL). [Yes]
   \item All the training details (e.g., data splits, hyperparameters, how they were chosen). [Yes]
         \item A clear definition of the specific measure or statistics and error bars (e.g., with respect to the random seed after running experiments multiple times). [Yes]
         \item A description of the computing infrastructure used. (e.g., type of GPUs, internal cluster, or cloud provider). [Yes]
 \end{enumerate}

 \item If you are using existing assets (e.g., code, data, models) or curating/releasing new assets, check if you include:
 \begin{enumerate}
   \item Citations of the creator If your work uses existing assets. [Yes]
   \item The license information of the assets, if applicable. [Not Applicable]
   \item New assets either in the supplemental material or as a URL, if applicable. [Not Applicable]
   \item Information about consent from data providers/curators. [Not Applicable]
   \item Discussion of sensible content if applicable, e.g., personally identifiable information or offensive content. [Not Applicable]
 \end{enumerate}

 \item If you used crowdsourcing or conducted research with human subjects, check if you include:
 \begin{enumerate}
   \item The full text of instructions given to participants and screenshots. [Not Applicable]
   \item Descriptions of potential participant risks, with links to Institutional Review Board (IRB) approvals if applicable. [Not Applicable]
   \item The estimated hourly wage paid to participants and the total amount spent on participant compensation. [Not Applicable]
 \end{enumerate}

 \end{enumerate}

\appendix
\newpage
\onecolumn
\addcontentsline{toc}{section}{Appendix} % Add the appendix text to the document TOC
\part{Appendix} % Start the appendix part
\parttoc % Insert the appendix TOC

\section{Proof of Proposition~\ref{prop:lp}} \label{sec:proof}
\begin{proof}[Proof of Proposition~\ref{prop:lp}]
    Note that the constraint $\lvert \textbf{w}\rvert_0 \le n$ is automatically satisfied when we use the simplex method or its variant.
    Without this constraint, we have a trivial feasible solution $\textbf{w}=\textbf{w}_\text{cand}$,
    so, for the optimal solution $\textbf{w}_*$, we have
    $\textbf{w}_*^\top \bigl[ g(\textbf{X}_\text{cand}) \odot q(\textbf{X}_\text{cand})\bigr]
        \ge \textbf{w}_\text{cand}^\top \bigl[ g(\textbf{X}_\text{cand}) \odot q(\textbf{X}_\text{cand})\bigr]$.
    Since $\mathbb{E}[\tilde{\textbf{w}}_\text{batch}^\top g(\tilde{\textbf{X}}_\text{batch})] = \textbf{w}_\text{batch}^\top \bigl[ g(\textbf{X}_\text{batch}) \odot q(\textbf{X}_\text{batch})\bigr]
    = \textbf{w}_*^\top \bigl[ g(\textbf{X}_\text{cand}) \odot q(\textbf{X}_\text{cand})\bigr]$,
    we obtain the first estimate Eq.~(\ref{eq:lp-1}).

    For the latter estimate, we first decompose the error into
    two parts:
    \begin{align}
        &\mathbb{E}\!\left[\left\lvert \tilde{\textbf{w}}_\text{batch}^\top
        f(\tilde{\textbf{X}}_\text{batch}) -
        \textbf{w}_\text{cand}^\top f(\textbf{X}_\text{cand})\right\rvert\right]
        \nonumber\\
        &\le
        \mathbb{E}\!\left[\left\lvert \tilde{\textbf{w}}_\text{batch}^\top
        f(\tilde{\textbf{X}}_\text{batch}) -
        \textbf{w}_\text{batch}^\top f(\textbf{X}_\text{batch})\right\rvert\right]
        +
        \left\lvert \textbf{w}_\text{batch}^\top
        f(\textbf{X}_\text{batch}) -
        \textbf{w}_\text{cand}^\top f(\textbf{X}_\text{cand})\right\rvert
        \label{eq:dcp}.
    \end{align}
    For the first term, considering each $x\in \textbf{X}_\text{batch}$
    on whether or not it gets included in $\tilde{\textbf{X}}_\text{batch}$,
    we have
    \begin{align*}
        &\mathbb{E}\!\left[\left\lvert \tilde{\textbf{w}}_\text{batch}^\top
        f(\tilde{\textbf{X}}_\text{batch}) -
        \textbf{w}_\text{batch}^\top f(\textbf{X}_\text{batch})\right\rvert\right]
        \\&\le \textbf{w}_\text{batch}^\top
        \Bigl[ \lvert f\rvert(\textbf{X}_\text{batch}) \odot (1-q)(\textbf{X}_\text{batch})\Bigr]
        \le \textbf{w}_\text{batch}^\top (1-q)(\textbf{X}_\text{batch})
        \max_{x\in\textbf{X}_\text{batch}} \lvert f(x)\rvert
        \\&=\Bigl[1 - \textbf{w}_\text{batch}^\top q(\textbf{X}_\text{batch})\Bigr]
        \max_{x\in\textbf{X}_\text{batch}} \lvert f(x) \rvert
        \le \Bigl[1 - \textbf{w}_\text{cand}^\top q(\textbf{X}_\text{cand})\Bigr]
        \max_{x\in\textbf{X}_\text{batch}} \lvert f(x) \rvert,
    \end{align*}
    where the last inequality follows from the inequality constraint
    $(\textbf{w} - \textbf{w}_\text{cand})^\top q(\textbf{X}_\text{cand}) \ge 0$
    in the LP.
    Since $\lvert f(x)\rvert=\lvert\langle f, K_\text{LP}(\cdot, x) \rangle\rvert
    \le \lVert f\rVert K_\text{LP}(x, x)^{1/2}$ from the reproducing property of RKHS,
    we obtain
    \begin{equation}
        \mathbb{E}\!\left[\left\lvert \tilde{\textbf{w}}_\text{batch}^\top
        f(\tilde{\textbf{X}}_\text{batch}) -
        \textbf{w}_\text{batch}^\top f(\textbf{X}_\text{batch})\right\rvert\right]
        \le \epsilon_\text{rej}K_{\max}\lVert f\rVert.
        \label{eq:proof-lp-1}
    \end{equation}

    Let us then bound the second term of the RHS of Eq.~(\ref{eq:dcp}).
    Note that, from the formula of worst-case error of kernel quadrature
    (see, e.g., \citep[][Eq. (14)]{hayakawa2022positively}),
    we can bound
    \begin{align}
        \left\lvert \textbf{w}_\text{batch}^\top
        f(\textbf{X}_\text{batch}) -
        \textbf{w}_\text{cand}^\top f(\textbf{X}_\text{cand})\right\rvert^2
        \le \lVert f \rVert^2
        (\textbf{w}_* - \textbf{w}_\text{cand})^\top
        K_\text{LP}(\textbf{X}_\text{cand}, \textbf{X}_\text{cand}) (\textbf{w}_* - \textbf{w}_\text{cand})
        \label{eq:proof-lp-2}
    \end{align}
    (recall $\textbf{w}_*$ has the same dimension as $\textbf{w}_\text{cand}$).
    We now want to estimate
    \begin{equation}
        (\textbf{w}_* - \textbf{w}_\text{cand})^\top
        K_\text{LP}(\textbf{X}_\text{cand}, \textbf{X}_\text{cand}) (\textbf{w}_* - \textbf{w}_\text{cand}).
        % \le (2\epsilon_\text{nys} + \epsilon_\text{LP})^2.
        \nonumber
    \end{equation}
    Consider approximating $K_\text{LP}$ by $K_\text{nys}$.
    Since $K_\text{LP} - K_\text{nys}$ is positive semi-definite from the property
    of Nystr{\"o}m approximation (see, e.g., the proof of \citep[][Corollary 4]{hayakawa2022positively}),
    for any $x, y \in \textbf{X}_\text{cand}$, we have
    \[
        \lvert (K_\text{LP} - K_\text{nys})(x, y)\rvert
        \le \lvert (K_\text{LP} - K_\text{nys})(x, x)\rvert^{1/2}
        \lvert (K_\text{LP} - K_\text{nys})(y, y)\rvert^{1/2}
        \le \epsilon_\text{nys}^2.
    \]
    Thus, we have
    \begin{align}
        &(\textbf{w}_* - \textbf{w}_\text{cand})^\top
        \Bigl[(K_\text{LP} - K_\text{nys})(\textbf{X}_\text{cand}, \textbf{X}_\text{cand})\Bigr] (\textbf{w}_* - \textbf{w}_\text{cand})
        \nonumber\\
        &\le (\textbf{w}_* + \textbf{w}_\text{cand})^\top (\epsilon_\text{nys}^2\boldsymbol{1}\boldsymbol{1}^\top) (\textbf{w}_* + \textbf{w}_\text{cand}) = 4\epsilon_\text{nys}^2.
        \label{eq:proof-lp-3}
    \end{align}
    Finally, we estimate
    \begin{align}
        &(\textbf{w}_* - \textbf{w}_\text{cand})^\top
        K_\text{nys}(\textbf{X}_\text{cand}, \textbf{X}_\text{cand}) (\textbf{w}_* - \textbf{w}_\text{cand})
        \nonumber \\
        &=(\textbf{w}_* - \textbf{w}_\text{cand})^\top
        \sum_{j=1}^{n-2}
        \boldsymbol{1}_{\{\lambda_j>0\}}\lambda_j^{-1}
        \varphi_j(\textbf{X}_\text{cand})\varphi_j(\textbf{X}_\text{cand})^\top
        (\textbf{w}_* - \textbf{w}_\text{cand}) \nonumber\\
        &=\sum_{j=1}^{n-2}
        \boldsymbol{1}_{\{\lambda_j>0\}}\lambda_j^{-1}
        \left[(\textbf{w}_* - \textbf{w}_\text{cand})^\top \varphi_j(\textbf{X}_\text{cand})\right]^2.
        \label{eq:proof-lp-4}
    \end{align}
    From the inequality constraint in the LP,
    we have $\lvert(\textbf{w}_* - \textbf{w}_\text{cand})^\top \varphi_j(\textbf{X}_\text{cand})\rvert \le \epsilon_\text{LP}
    \sqrt{\lambda_j/(n-2)}$,
    so that Eq.~(\ref{eq:proof-lp-4}) is further bounded as
    \begin{align}
        (\textbf{w}_* - \textbf{w}_\text{cand})^\top
        K_\text{nys}(\textbf{X}_\text{cand}, \textbf{X}_\text{cand}) (\textbf{w}_* - \textbf{w}_\text{cand}) \le \sum_{j=1}^{n-2}
        \boldsymbol{1}_{\{\lambda_j>0\}}\lambda_j^{-1}
        \epsilon_\text{LP}^2\frac{\lambda_j}{n-2} \le \epsilon_\text{LP}^2.
        \label{eq:proof-lp-5}
    \end{align}
    By adding the both sides of Eqs.~(\ref{eq:proof-lp-3}) and (\ref{eq:proof-lp-5}),
    we obtain
    \[
        (\textbf{w}_* - \textbf{w}_\text{cand})^\top
        K_\text{LP}(\textbf{X}_\text{cand}, \textbf{X}_\text{cand}) (\textbf{w}_* - \textbf{w}_\text{cand})
        \le 4\epsilon_\text{nys}^2 + \epsilon_\text{LP}^2
        \le (2\epsilon_\text{nys} + \epsilon_\text{LP})^2.
    \]
    By applying this to Eq.~(\ref{eq:proof-lp-2}), we have
    $\left\lvert \textbf{w}_\text{batch}^\top
        f(\textbf{X}_\text{batch}) -
        \textbf{w}_\text{cand}^\top f(\textbf{X}_\text{cand})\right\rvert
        \le \lVert f \rVert (2\epsilon_\text{nys} + \epsilon_\text{LP})$.
    Combining this with Eqs.~(\ref{eq:dcp}) and (\ref{eq:proof-lp-1}) yields
    the desired inequality Eq.~(\ref{eq:lp-2}).
\end{proof}

\section{Background}
\subsection{Gaussian process}\label{sup:gp}
GP \citep{rasmussen2006gaussian} is a widely used Bayesian regression model and the most popular surrogate model in PN. We consider the probabilistic model $\mathbb{P}(\boldsymbol{y}|\boldsymbol{x}, \theta)$ parameterized by $\theta \in \Theta$, mapping from inputs $x \in \mathcal{X}$ to a distribution over outputs/labels $y \in \mathcal{Y}$. Here, the labels can potentially only be observed through a noisy estimate, $y = f(x) + \epsilon$, where $y$ is a continuous value and the noise $\epsilon \sim \mathcal{N}(0, \lambda^2)$ is assumed to be generated by i.i.d. zero-mean Gaussian, and $\lambda^2$ is the noise variance.
Given a labelled dataset $\mathcal{D}_0 = \{ \boldsymbol{x}_n, \boldsymbol{y}_n\}_{n=1}^N := (\mathcal{X}_N, \mathcal{Y}_N)$, GP regression model is given by $f\mid D_0 \sim \mathcal{GP}(m, C)$, where
\begin{align}
    \begin{split}
    m(x) &= K(x, \boldsymbol{x}_0) \boldsymbol{K}_\lambda^{-1} \boldsymbol{y}_0,\\
    C(x, x^\prime) &= K(x, x^\prime) - K(x, \boldsymbol{x}_0) \boldsymbol{K}_\lambda^{-1} K(\boldsymbol{x}_0, x^\prime),
    \end{split}\label{eq:gp}
\end{align}
$f$ is the surrogate function, $m(\cdot)$ and $C(\cdot, \cdot)$ are the mean and covariance of posterior predictive distribution of $f$, and $\mathcal{D}_0 = (\boldsymbol{x}_0, \boldsymbol{y}_0)$ is the observed dataset. %of the model parameter $\boldsymbol{x}_0$ and corresponding likelihood value $\boldsymbol{y}_0:= f(\boldsymbol{x}_0) + \epsilon$, 
$K$ is the kernel parameterized by $\theta$\footnote{We typically assume zero mean GP prior over function space $\mathcal{GP}(0, K)$, and we assume Gaussian likelihood $\mathcal{N}(0, \lambda^2)$, hence the resulting posterior distribution is closed-form as shown in Eq.~\labelcref{eq:gp}, thanks to Gaussianity. Throughout the paper, we refer to a symmetric positive semi-definite kernel just as a kernel.}
and $\boldsymbol{K}_\lambda^{-1} := [K(\boldsymbol{x}_0, \boldsymbol{x}_0) + \lambda^2 \boldsymbol{I}]^{-1}$, and $\lambda^2 \boldsymbol{I}$ is the diagonal likelihood variance matrix.

\subsection{Fully Bayesian Gaussian Process}\label{sup:fbgp}
In this paper, we will consider the Bayesian active learning for Fully Bayesian Gaussian Process (FBGP) \citep{riis2022bayesian}. FBGP extends a GP by placing a prior over the hyperparameter $\mathbb{P}(\theta)$ and approximating their full posteriors. The predictive posterior for the test inputs $x^*$ is
\begin{align*}
\begin{split}
&\mathbb{P}(y^* \mid x^*, \mathcal{D}_0)\\
&= \iint \mathbb{P}(y^* \mid f^*, \theta,  x^*, \mathcal{D}_0) \text{d} \mathbb{P}(f^* \mid \theta,  x^*, \mathcal{D}_0) \text{d} \mathbb{P}(\theta \mid \mathcal{D}_0).
\end{split}
\end{align*}
While the inner integral for $f$ reduces to the normal GP predictive posterior, the outer integral for $\theta$ remains intractable and typically approximated by MCMC.

\subsection{Bayesian Quadrature}\label{sup:bq}
Bayesian quadrature (BQ) is an algorithm for evaluating integrals given by:
\begin{equation}
    \hat{Z} = \int_\mathcal{X} f(x) \text{d}\pi(x), \label{eq:bq}
\end{equation}
where $f$ is the black-box function we wish to integrate against a known probability measure $\pi$.
The difference from BO is the objective being integration, not global optimisation. The integration problem is widely recognised in statistical learning: expectations, variances, marginalisation, ensembles, Bayesian model selection, and Bayesian model averaging. BQ is, like BO, solved by GP-surrogate-model-based active learning. The batch acquisition methods are also shared with batch BO. The methodological differences are:
\begin{compactenum}
    \item BQ typically assumes a specific kernel to make the integration analytical (e.g.~RBF kernel).
    \item While BO requires to approximate the black-box function only in the vicinity of the global optimum, BQ needs to approximate the whole region of interest defined by the probability measure $\pi$.
\end{compactenum}
Thus, BQ is a purely explorative algorithm, and the uncertainty sampling acquisition function is often applied.

The classic method to estimate the integral exploits Gaussianity. Let $\pi$ be multivariate normal distribution $\pi(x) = \mathcal{N}(x; \mu_\pi, \boldsymbol\Sigma_\pi)$, and the kernel $K$ be RBF kernel, which can be represented as Gaussian $K(\boldsymbol{x}_0, x) = v \sqrt{|2\pi \textbf{W}|} \mathcal{N}(\boldsymbol{x}_0; x, \textbf{W})$, where $v$ is kernel variance and $\textbf{W}$ is the diagonal covariance matrix whose diagonal elements are the lengthscales of each dimension. As the product of two Gaussians is a Gaussian, the integrand becomes a Gaussian and its integral has the closed form, as such:
\begin{align}
\int m(x)\pi(x)dx 
&= v \left[ \int \mathcal{N}(x; \boldsymbol{x}_0, \textbf{W}) \mathcal{N}(x; \mu_\pi, \boldsymbol\Sigma_\pi) dx \right]^\top
\boldsymbol{K}^{-1}_\lambda \boldsymbol{y}_0,\\ 
&= v \left[ \int \mathcal{N}(x; \boldsymbol{x}_0, \textbf{W}) \mathcal{N}(x; \mu_\pi, \boldsymbol\Sigma_\pi) dx \right]^\top \boldsymbol{K}^{-1}_\lambda \boldsymbol{y}_0,\\ 
&= v \mathcal{N}(\boldsymbol{x}_0; \mu_\pi, \textbf{W}+\boldsymbol\Sigma_\pi)^\top \boldsymbol{K}^{-1}_\lambda \boldsymbol{y}_0,\\
&= \boldsymbol{z}^\top \boldsymbol{K}^{-1}_\lambda \boldsymbol{y}_0
\end{align}
where $\boldsymbol{z} := v \mathcal{N}(\boldsymbol{x}_0; \mu_\pi, \textbf{W}+\boldsymbol\Sigma_\pi)$. As such, the integration of GP over the measure $\pi$ is analytical. This $\boldsymbol{z}$ corresponds to the kernel mean in Eqs.~(\ref{eq:bq_mean})-(\ref{eq:bq_var}). This clearly explains that we need an analytical kernel mean to perform BQ. Thus, classical BQ methods have limitations on prior and kernel selections to be analytical. To make the integration closed-form, the prior needs to be uniform or Gaussian, and the kernel also needs to be limited selection (e.g. RBF kernel, see Table 1 in \cite{briol2019probabilistic}). Recent work \citep{adachi2022fast} extends this to arbitrary kernel and prior.

\section{Related Work}\label{sup:related}
\paragraph{Batch Active Learning}
A wide variety of batch methods has been proposed for each task: batch AL \citep{pinsler2019bayesian, kirsch2019batchbald, riis2022bayesian}, batch BQ \citep{wagstaff2018batch, adachi2022fast, adachi2023bayesian} and batch BO, a greedy extension of sequential algorithms \citep{azimi2010batch, gonzalez2016batch, eriksson2019scalable, balandat2020botorch}, diversified batch with determinantal point process (DPP) \citep{kathuria2016batched, nava2022diversified}. Constrained batch construction has been researched in BO community \citep{hernandez2016a, benjamin2019constraned, eriksson2021scalable}. However, most works do not discuss the relationship to quality of batch construction like KQ methods.

\paragraph{Kernel Quadrature}
For general KQ methods, There are a number of KQ algorithms; herding/optimization \citep{chen2010super, bach2012on, huszar2012optimally}, random sampling \citep{bach2017on, belhadji2019kernel}, DPP \citep{belhadji2019kernel, belhadji2021analysis}, kernel thinning \citep{dwivedi2021kernel, dwivedi2022generalized}, recombination \citep{hayakawa2022positively, hayakawa2023sampling}, kernel Stein discrepancy \citep{chen2018stein, chen2019stein, teymur2021optimal}, randomly pivoted Cholesky \citep{epperly2023kernel}. Similarly, Bayesian coresets is one of applications of quantization method and proposes a variety of algorithms \citep{campbell2019automated, manousakas2020bayesian, zhang2021bayesian, chen2022bayesian}, thus they are strongly related to KQ. KQ is a more proper framework for GP-based AL as it can incorporate the model uncertainty information for batch construction. 

\paragraph{Adaptive Batch Size}
While all of the above KQ/Bayesian coresets methods can be used for batch construction, almost all methods assume the batch size is predefined.
The adaptive batch size setting remains largely unsolved. In batch BO, \citet{nguyen2016budgeted} firstly formulated this setting as a Gaussian mixture fitting to the acquisition function and estimated the batch size as Bayesian model selection, which is obviously non-KQ-based. No other work proposes dynamic batch-size AL, to the best of our knowledge.

\paragraph{Connection to Bayesian Coresets}
While Bayesian coresets use Kullback-Leibler or Wasserstein divergence as a metric \citep{kim2022divergence} and weighted Euclidean inner product, KQ uses MMD. MMD and KQ have a direct relationship and KQ can incorporate additional information on model uncertainty to quantize the probability distribution. 
In KQ, the selected batch points are chosen to reduce the model uncertainty, which is advantageous property for active learning that needs to train model effectively.
Thus, KQ can utilize more information if the model has the analytical predictive covariance like GP. 
The quality of batch construction can be evaluated as the worst-case error in Eq.~\labelcref{eq:wce}, as this is equivalent to the MMD, which evaluates the divergence between quantized probability measure and the target distribution.

\section{Batch Bayesian Active Learning as Quantization}
\subsection{Sparse Subset Approximation}\label{sup:acs}
\citet{pinsler2019bayesian} proposed the batch construction heuristics with sparse subset approximation. The original paper did not state this but the formulaton is exactly the same as the weighted quantization task. They interpret this as Bayesian coresets \citep{campbell2019automated}, which is another view of weighted quantization and essentially close to KQ.
Their attempt is simple: they construct batch samples to best approximate the true posterior $\mathbb{P}(\theta | \mathcal{D}_0 \cup \mathcal{D}_t) \approx \mathbb{P}(\theta | \mathcal{D}_0 \cup \mathcal{D}_N)$. Here, the true posterior $\mathbb{P}(\theta | \mathcal{D}_0 \cup \mathcal{D}_N)$ means the posterior with the complete dataset $\mathcal{D}_N = (\mathcal{X}_N, \mathcal{Y}_N)$. Unsurprisingly, this true posterior is not available in the AL setting. While an unlabelled pool of candidates $\mathcal{X}_N$ is given, $\mathcal{Y}_N$ is not given. $\mathcal{Y}_N$ means the exhaustive number of costly human labelling, which we wish to reduce and is the motivation to perform AL. Hence, we have to approximate the true posterior. \citet{pinsler2019bayesian} approximated the true posterior using the expectation of the current posterior with respect to the current predictive distribution.
\begin{align*}
    \mathbb{P}(\theta \mid \mathcal{X}_N, \mathcal{D}_0) = \int \mathbb{P}(\theta \mid \mathcal{D}_0, \mathcal{X}_N, y^*) \text{d} \mathbb{P}(y^* \mid \mathcal{D}_0, \mathcal{X}_N),
\end{align*}
where $\mathbb{P}(y^* \mid \mathcal{D}_0, \mathcal{X}_N)$ is the predictive posterior for the unlabelled inputs $\mathcal{X}_N$. This \textit{expected} posterior is different from the current posterior $\mathbb{P}(\theta \mid \mathcal{D}_0)$ as it is explicitly conditioned on $\mathcal{X}_N$. In other words, the current posterior is not conditioned on the input $x$, thus it is not useful to guide the next query $\mathcal{X}_t$. This expected posterior is conditioned on the input, hence this can guide the next query points.
They used this expected posterior as the target distribution in the quantization task, then constructed the batch samples using Bayesian coresets. Namely, approximating expected posterior using the subset $\mathcal{X}_t \subset \mathcal{X}_N$, $\mathbb{P}(\theta \mid \mathcal{X}_N, \mathcal{D}_0) \approx \mathbb{P}(\theta \mid \mathcal{X}_t, \mathcal{D}_0)$.
This heuristic outperformed popular baselines, such as BALD \citep{houlsby2011bayesian} and clustering-based approach.

\subsection{Reinterpret as Kernel Quadrature}
We reinterpret this formulation as a KQ task.
\begin{compactenum}
    \item The target distribution $\mu$ is the weighted samples $(\textbf{w}_\text{cand}, \textbf{X}_\text{cand})$, where $\textbf{X}_\text{cand} := \mathcal{X}_N$ is the candidate samples (the unlabelled pool), and $\textbf{w}_\text{cand} \propto \mathbb{P}(\theta \mid x_l, \mathcal{D}_0)$ is the weights of candidate samples and $x_l \in \textbf{X}_\text{cand}$. The weights are the expected posterior as introduced in the section~\ref{sup:acs}.
    \item The quantized distribution $\nu$ is the weighted samples $(\textbf{w}_\text{batch}, \textbf{X}_\text{batch})$, and $\textbf{X}_\text{batch} \subset \textbf{X}_\text{cand}$ is the next batch query points.
    \item The kernel is the \emph{expected} predictive covariance of FBGP model, $K(\cdot, \cdot) := \mathbb{E}_{\theta \sim \mathbb{P}(\theta \mid \mathcal{D}_0)}[C(\cdot, \cdot \mid \theta)]$. This enables us to incorporate the model uncertainty to construct the quantized samples unlike the original paper \citep{pinsler2019bayesian}. The original work used the weighted Euclidean inner product using only the expected posterior, which loses the information of uncertainty.
\end{compactenum}
As such, we can reinterpret the sparse subset approximation method as a KQ task. Hence, we can apply our LP formulation in the AdaBatAL to batch AL tasks. Furthermore, we have additional room for incorporating the information in LP formulation; the reward $g$. We used the B-QBC acquisition function as the reward for incorporating the additional information on estimation variance in mean estimation.

\section{Batch Bayesian Optimization as Quantization}
\subsection{Primer of Bayesian Optimization}\label{sup:primer}
BO \citep{mockus1998application, garnett2023bayesian} aims to optimize the blackbox function $f$ when there is no access to the closed-form function nor gradient but can query the function pointwise.
\begin{align}
    x^*_\text{true} = \mathop\mathrm{argmax}_{x \in \mathcal{X}} f(x), \label{eq:obj_bo}
\end{align}
where $x^*_\text{true}$ is the ground truth of the global optimum. We wish to find as large $f(x)$ as possible under some given budget, such as the overall cost or number of queries. BO is a surrogate-model-based optimizer, which typically adopts GP. BO is also extended to batch BO. The core difference between BO and BQ is that BO only needs an accurate model in the vicinity of global optimal locations, whereas BQ needs an accurate model all over the domain. Therefore, BQ can be viewed as a pure exploration algorithm (as BQ explores based on only uncertainty, as shown in Eq.~\labelcref{eq:kq-eq}).

\subsection{Batch Bayesian Optimization as a Kernel Quadrature}\label{sup:sober}
\citet{adachi2023sober} has introduced the heuristics to recast the batch BO as a KQ task:
\begin{align}
    \delta_{x^*_\text{true}} \in \mathop\mathrm{argmax}_{\pi} \int f(x) \text{d}\pi(x),  \label{eq:dual}
\end{align}
where $\delta_x$ is the delta distribution at $x$ and $\pi(x) := \mathbb{P}(f(x) = \max_{x \in \mathcal{X}} f(x))$ is a probability distributions (belief) of $x^*_\text{true}$. Note that the optimization target in Eq.~\labelcref{eq:dual} has now changed from $x$ into $\pi$. We view Eq.~\labelcref{eq:dual} as a batch-sequential `measure' ($\pi$) optimization, which updates $\pi$ over each iteration. The more data we observe, the more confidently we can estimate the location of the global maximum. This corresponds to that the distribution $\pi$ `shrinks' toward the true global optimum location, and becomes the delta distribution in the ideal case of a single global maximum. The role of $\pi$ can be understood as `exploitation', which determines the promising region over the domain.

The intuition of this reformulation is as follows:
\begin{compactenum}
    \item[(a)] Measure optimization is \textit{dual} to original global optimization \citep{lasserre2011new}. Classically, a deterministic polynomial regressor \citep{lasserre2011new, martinez2020quadrature, rudi2020finding} has been applied with provable convergence rate.
    \item[(b)] Measure optimization is \textit{convex optimization} with a linear objective function even if the function $f$ is non-convex \citep{rudi2020finding}.
    \item[(c)] Convexity negates the necessity to maximize the non-convex multimodal acquisition function. Thus, it provides a computationally efficient solution without worrying acquisition function being properly maximised at each iteration and batch in typical batch heuristics.
    \item[(d)] Variance of $\pi$ correlates with the variance of predictive distribution under $\pi$ \citep{adachi2023sober}. Thus, minimising the variance of $\pi$ leads to minimising the GP predictive variance under $\pi$, which is exactly the BQ task and finding the batch points is exactly the KQ task via the duality in Eq.~\labelcref{eq:kq-eq}
    \item[(e)] Unlike typical BO, BQ has the robustness guarantee when RKHS is misspecified \citep{kanagawa2016convergence, hayakawa2022positively}. This is advantageous as GP in BO tends to be suboptimally tuned \citep{ha2023provably}.
\end{compactenum}
This heuristic approach showed the state-of-the-art performance as batch BO over commonly used 8 heuristics of batch BO\citep{adachi2023sober}.
Similar to the fact that many acquisition functions have been proposed, a variety of $\pi$ definitions can be adopted. \citet{adachi2023sober} has proposed two variants: (i) Thompson sampling (TS) \citep{thompson1933likelihood}, $\pi:= \mathbb{P}(x^* | \boldsymbol{D}_t)$, where $x^*:= \mathop\mathrm{argmax}_x f$ is the maximum location of $f$, a sample from GP predictive posterior distribution, $\boldsymbol{D}_t:= (\boldsymbol{x}_\text{obs, t}, \boldsymbol{y}_\text{obs, t})$ is the dataset we observed until $t$-th iterations. That is, we use the current belief of the maximum location of surrogate model $x^*$ instead of ground truth $x^*_\text{true}$. (ii) Probability of improvement (PI) \citep{kushner1964new}: $\pi:= \mathbb{P}(f \geq \eta | \boldsymbol{D}_t)$, where $\eta := \mathop\mathrm{argmax}_x \textbf{D}_t$. In both cases, $\pi$ shrinks toward $x^*_\text{true}$ upon $\boldsymbol{D}_t$ updates.

\subsection{Defining as Kernel Quadrature}
We define this formulation as a KQ task.
\begin{compactenum}
    \item The target distribution $\mu$ is the weighted samples $(\textbf{w}_\text{cand}, \textbf{X}_\text{cand})$, where $\textbf{X}_\text{cand} \sim \pi(x)$ is the candidate samples drawn from the probability distribution of $x_\text{true}$.  $\textbf{w}_\text{cand}$ can be equal weights or importance weights when proposal distribution is applied \citep{adachi2023sober}.
    \item The quantized distribution $\nu$ is the weighted samples $(\textbf{w}_\text{batch}, \textbf{X}_\text{batch})$, and $\textbf{X}_\text{batch} \subset \textbf{X}_\text{cand}$ is the next batch query points.
    \item The kernel is the predictive covariance of normal GP, $K(\cdot, \cdot) := C(\cdot, \cdot \mid \theta)$.
\end{compactenum}
As such, regardless of the $\pi$ definition, batch BO can be solved as a KQ task.

\subsection{Constrained Bayesian Active Learning and Optimization}\label{sup:cBO}
Suppose we have constraints with which we must comply, but we do not know the constraint functions a priori and are only observable pointwise. We can model such constraint functions by GPs, similarly to the surrogate model of the objective function. We can classify the type of constraints into (A) continuous, and (B) binary constraints.

\subsubsection{Modelling Continuous Constraints}
Continuous constraints naturally appear in the form of threshold (e.g. controlling a car not to exceed the speed limit, or maximize the computer power not to exceed the temperature limit), given by:
\begin{equation}
    Q_\ell(x) \geq 0 \label{eq:contconst}
\end{equation}
where $Q_\ell$ is the $\ell$-th latent constraint function. We can reformulate most constraints to be the form of Eq.~(\ref{eq:contconst}). For instance, when we wish for the temperature not to surpass the limit $T_\text{limit}  \geq T$, we can set $Q_\ell(T) = T_\text{limit} - T$.
We assume we can query these latent values $g_\ell(x)$ at the designated location $x$, but the function itself is unknown. We need to guess the function shape only from queries. 

We place a GP regression model on the latent values $Q_\ell(x)$. Then, the probability of constraint satisfaction $q_\ell$ can be given:
\begin{equation}
    q_\ell(x) := \mathbb{P}(Q_\ell(x) \geq 0) = \Phi \left( \frac{m_\ell(x)}{\sqrt{C_\ell(x, x)}} \right)
\end{equation}
where $m_\ell$ and $C_\ell$ are the posterior predictive mean and covariance of GP on the $\ell$-th constraint, $\Phi(x)$ is the cumulative distribution function of the standard normal distribution $\mathcal{N}(x; 0, 1)$.

\subsubsection{Modelling Binary Constraints}
Binary constraints return the constraint satisfaction as a Boolean value (yes or no). This is typically modelled with a GP classifier: As the classification likelihood such as Bernoulli likelihood is not conjugate with GP prior, the resulting posterior predictive distribution is no more closed-form. As such, we normally estimate the posterior predictive distribution via sampling functions from latent space, then transform them via the so-called link function \cite{rasmussen2006gaussian}.

We adopted an approach with Dirichlet-based GP (DGP) \citep{milios2018dirichlet} for scalability. Let $f_\ell \sim \mathcal{GP}(m_\ell, C_\ell)$ be the GP classifier modelling the $\ell$-th binary constraint, the binary feedback $y = 1$ be the constraint satisfaction, $y = 0$ be the constraint violation.
Monte Carlo integration via transforming the sampled function with the link function can estimate the expectation of binary probability as Bernoulli distribution:
\begin{align}
    q_\ell(x) := \mathbb{P}(y=1 \mid x) = \int \frac{\exp(f_{\ell, i})}{\sum \exp(f_{\ell, i})} \mathbb{P}(f_{\ell, i} | x, \textbf{D}_\ell) \text{d} f_\ell
\end{align}
where $\textbf{D}_\ell$ is the observed dataset of the constraint satisfaction $\textbf{y}_\ell$ at the inputs $\textbf{x}_\ell$.

\subsubsection{Constrainted Active learning and Optimization}\label{sup:constraint}
With the above constraint models, the typical constrained BO is performed by constraining the acquisition function, namely, $\alpha(x) \prod_{\ell=1}^c q_\ell(x)$ \citep{gardner2014bayesian, gelbart2014bayesian}. In an AdaBatAl algorithm, we can incorporate the constraint information as $q$ in the LP formulation. We do not need to modify the acquisition function.

\section{Kernel Quadrature for Intractable Kernel Mean}\label{sup:empiricalmeasure}
\begin{table}[htb!]
\centering
\caption{How to set the target distribution for each active learning task.}
\label{tab:task_diff}
\begin{tabular}{lll}
\toprule
 task & target distribution & meaning \\
\midrule
 Active learning & $\mathcal{X}_N$ & unlabelled pool of candidate inputs \\
 Bayesian optimization & $\mathbb{P}(f(x) = \max_{x \in \mathcal{X}} f(x))$  & probability distribution of global optimum location \\
 Bayesian quadrature  & $\mathbb{P}(x)$ & prior distribution\\
\bottomrule
\end{tabular}
\end{table}
Table~\ref{tab:task_diff} summarizes the target distribution definitions for each AL, BO, and BQ task. While active learning considers the discrete candidates, BO and BQ consider continuous distributions. As seen in the section \ref{sup:bq}, only a handful of combinations of continuous target distributions and the kernels can provide the analytical kernel mean and variance, thereby providing the analytical expectation of test functions in the section~\ref{sec:nystrom}. This is not always true, particularly for intractable probability distribution (e.g. Thompson sampling in BO), and/or intractable kernel (e.g. Tanimoto kernel for drug discovery tasks). We review how to construct a KQ task for such an intractable pair of target distribution and kernel in this section.

\paragraph{Intractable Expectations of Test Functions}
We consider \textit{approximately} constructing the probability measure to estimate the expectation. We construct \textit{empirical measure} $\mu_\text{cand}(x) := \sum_{i=1}^P w_i \delta_{x_i} := (\textbf{w}_\text{cand}, \textbf{X}_\text{cand})$, where $\textbf{X}_\text{cand} \subset \textbf{X}^N$. We draw very large $N$ samples to approximate the expectations, which assumes $N$ is sufficiently larger than the batch size $n$, $N \gg n$. Hence, we can approximate $\int_\mathcal{X} \boldsymbol\varphi(x) \text{d} \mu(x) \approx \textbf{w}_\text{cand}^T \boldsymbol\varphi(\textbf{X}_\text{cand})$. This permits kernel quadrature for any pair $(K, \mu)$, unlike the original BQ.
If directly sampling from $\mu(x)$ is expensive, the empirical measure can be constructed with importance weights. Namely, we draw large samples from cheaper-to-sample distribution (e.g., domain $\mathcal{X}$), then calculate the importance weights by taking the ratio of the probability density function (see details in \citep{adachi2023sober}). 

\paragraph{Error Bounds}
While the empirical measure makes BQ/KQ applicable to an arbitrary combination of $(K, \mu)$, this produces an additional approximation error.
The total error bounds of this KQ method are given by:
\begin{align}
\begin{split}
    \text{wce}(Q_{\boldsymbol{w}})
    \leq 2 \sup_{x, y} \sqrt{K(x,y) - K_0(x, y)} + \text{MMD}_\mathcal{H}(\mu, \mu_\text{cand}).
\end{split}\label{eq:rate}
\end{align}
The proof is given in Proposition 1 in \citep{hayakawa2022positively}. The first and second terms correspond to the Nyström and the empirical measure approximation error, respectively. When we take large $M$ Nyström samples and $N$ empirical measures, we can make error bounds tighter if the time budget allows. In practice, we take $N=20,000$ and $M=500$ for the batch size $n \leq 100$. (see Appendix in \citep{adachi2022fast} for how to set these values).

\section{Experimental Details}\label{sup:exp}
\subsection{Training details}
We have tested AdaBatAL for 7 synthetics and 7 real-world tasks for batch AL and BO tasks. Our experiments were repeated 10 times and took a mean and one standard error with different random seeds (the seeds are shared with baseline methods). We use FBGP for batch AL tasks, and simple GP with type-II maximum likelihood estimation for batch BO tasks. The kernel is different for each task but shared with baseline methods (see details in the dataset section). We randomly generated 10 samples as the initial dataset $\mathcal{D}_0$. We use different batch sizes for each task (see details in the dataset section). While the fixed batch size methods simply adopt this as the batch size, AdaBatAL sets this as the upper bound of batch sizes. This means the AdaBatAL tends to query a smaller number of samples than fixed batch size methods.
We iterated this batch acquisition process for the fixed iteration times and compared the best-observed values at the last round. For the fair comparison with the adaptive batch size method, we employ the accumulated queries as the metric, which counts the total number of queries at the $t$-th iteration. As explained, AdaBatAL yields the smaller accumulated queries with the same iteration times. For constrained cases, we removed the violated samples. Thus, constrained tasks yield smaller accumulated queries than unconstrained cases even with the same batch sizes and the same iteration times. Surprisingly, non-adaptive batch baselines tend to have smaller batch sizes than adaptive AdaBatAL due to constraint violation (See Figure~\ref{fig:batchBO}).

Our code is built upon PyTorch-based libraries \citep{paszke2019pytorch, gardner2018gpytorch, balandat2020botorch, griffiths2022gauche} and Gurobipy \citep{gurobi} is used to solve the linear programming. All baseline methods are official implementations in BoTorch or coded with BoTorch \citep{balandat2020botorch}.

\paragraph{Batch Bayesian Optimization}
We use a constant-mean GP with either RBF, Tanimto, or graph diffusion kernel for batch BO tasks. In each iteration of the active learning loop, the outputs are standardized to have zero mean and unit variance. We optimize the hyperparameter by maximizing the marginal likelihood (type-II maximum likelihood estimation) using L-BFGS-B optimizer \citep{liu1989limited} implemented with BoTorch \citep{balandat2020botorch}. The initial data sets consist of ten data points drawn by Sobol sequence \citep{sobol1967distribution}, and in each iteration, multiple data points are queried as the batch acquisition (upper bound for AdaBatAL). 
We adopt log regret if the true global maxima are known, otherwise, the log of best-observed value is the evaluation metric using the test dataset. The models are implemented in GPyTorch \citep{gardner2018gpytorch}. All experiments are repeated ten times with different initial data sets via different random seeds.

\paragraph{Batch active learning}
We use a zero-mean GP with an RBF kernel for all batch AL tasks. In each iteration of the active learning loop, the inputs are rescaled to the unit cube $[0,1]^d$, and the outputs are standardized to have zero mean and unit variance. Following \citet{lalchand2020approximate}, we give all the hyperparameters relatively uninformative $\mathcal{N}(0, 3)$ lognormal priors. 
The initial data sets consist of ten data points drawn by Sobol sequence \citep{sobol1967distribution}, and in each iteration, 10 data points are queried as the batch acquisition (upper bound for AdaBatAL). The unlabeled pool consists of the 10,000 data points drawn by Sobol sequence all over the domain. We used this unlabelled pool and corresponding true values as the test dataset for the evaluation. We adopt negative log marginal likelihood (NLML) as the evaluation metric using the test dataset. The inference in FBGP is carried out using NUTS \citep{hoffman2014no} in Pyro \citep{bingham2019pyro} with five chains and 500 samples, including a warm-up period with 200 samples. The remaining 1500 samples are all used for the acquisition functions. The models are implemented in GPyTorch \citep{gardner2018gpytorch}. All experiments are repeated ten times with different initial data sets via different random seeds. 

For batch AL, we typically assume training a model is very expensive (e.g. deep learning). FBGP is expensive to train even with parallel chains. Thus, we exclude methods like hallucination that require the sequential update of the model to select multiple points. This assumption is widely shared with the AL community (e.g. \citet{kirsch2019batchbald, pinsler2019bayesian}). 
Moreover, all baseline batch AL methods do not consider probabilistic constraints. We simply follow the constrained BO approaches, explained in section~\ref{sup:constraint}.

\paragraph{Extension to non-continuous input domain}
Almost all methods are not compatible with categorical and mixed input spaces due to the continuity assumption in these methods. To enable comparison against these methods, we adopt the nearest neighbor in discrete or mixed problems: namely, we optimise the discrete variables as bounded continuous variables, then the selected continuous locations are classified into the closest original discrete values. For the graph space, we deem the search space itself to be a graph and the objective is to find a subgraph. This is different from, for example, the drug discovery problem, whose input variables are graphs but the space itself is a non-Euclidean discrete set of drugs. In contrast, the graph space is over the large graph, and the graph example is only one. Thus, cTS is the only method applicable to graph space other than AdaBatAL.

\paragraph{Extension to constrained cases}
We simply follow the constrained BO approaches, explained in the section~\ref{sup:constraint}; Modelling the probabilistic constraints by GPs and multiplying the probability of constraint satisfaction to the acquisition function. 

\paragraph{Training details of AdaBatAL}
For AdaBatAL, we have two hyperparameters; the number of Nyström samples $M$, and the tolerance $\epsilon_\text{LP}$. The number of unlabeled pools $N$, the batch sizes $n$, and $M$ need to satisfy the relationship $N \gg M \geq n$. We fixed $M=500$. As explained in the section~\ref{sup:empiricalmeasure}, the larger $M$ yields tighter error bounds for worst-case error but it slows down the computation. We find $M=500$ works well over the tasks we have tested. For $\epsilon_\text{LP}$, this is automatically determined for the constrained case via $\epsilon_\text{LP} = \epsilon_\text{vio}$. For unconstrained cases, we set $\epsilon_\text{LP} = 0.01$. For reward function $g$, we set B-QBC \citep{riis2022bayesian} for batch AL, and no reward function is set for batch BO. The probabilistic constraints $q$ were modeled by GP as explained. For the intractable expectation of kernel means, we generate $N=20,000$ data points from the probability distribution $\mu$ as explained in section~\ref{sup:empiricalmeasure}.

\subsection{Baseline Implementations}
\begin{table}[hbt!]
\centering
\caption{Summary of baseline method. cBO refers to constrained BO.}
\label{tab:baselines}
\begin{tabular}{llllllll}
\toprule
 method & task & adaptive? & constraints? & discrete? & large batch? & any kernel? & any AF? \\
\midrule
 random & any & & & \ding{51} & \ding{51} & \ding{51} & \ding{51} \\
 \textbf{AdaBatAL (ours)} & any & \ding{51} & \ding{51}& \ding{51} & \ding{51} & \ding{51} & \ding{51} \\
 \midrule
 B3O & BO & \ding{51} & &  & & \ding{51} & \ding{51} \\
 TS & BO &  & & \ding{51} & \ding{51} & \ding{51} & \\
 hallucination & BO & & & \ding{51} &  & \ding{51} & \ding{51} \\
 LP & BO &  & & & \ding{51} &  & \ding{51} \\
 TurBO & BO & & & & \ding{51} & & \ding{51} \\
 SOBER & BO & & & \ding{51} & \ding{51} & \ding{51} & \ding{51}\\
 \midrule
 MaxEnt & AL & & & \ding{51}& \ding{51} & \ding{51} & \\
 BALD & AL &  & & \ding{51} & \ding{51} & \ding{51} & \\
 B-QBC & AL & &  & \ding{51} & \ding{51} & \ding{51} & \\
 ACS-FW & AL & & & \ding{51} & \ding{51} & \ding{51} & \ding{51}\\
 \midrule
 cEI & cBO & & \ding{51} & & \ding{51} & \ding{51} & \\
 cTS & cBO & & \ding{51} & \ding{51} & \ding{51} & \ding{51} & \\
 SCBO & cBO & & \ding{51} & & \ding{51} &  & \\
 PropertyDAG & cBO & &\ding{51} & & \ding{51} & \\
 PESC & cBO & & \ding{51} & \ding{51} & & \ding{51} & \\
\bottomrule
\end{tabular}
\end{table}

Table~\ref{tab:baselines} summarizes all baselines. Our method, AdaBatAL, is the only method that can offer adaptive batch size under probabilistic constraints for both AL and BO tasks.

\subsubsection{Batch Bayesian Optimization}
\paragraph{B3O}
Budgeted Batch Bayesian Optimization (B3O) \citep{nguyen2016budgeted} is the only baseline method that offers the adaptive batch size. B3O recasts batch construction as the approximation of acquisition function using a mixture of Gaussians. The adaptive batch size is determined through the marginal likelihood of Gaussian mixture model; the number of Gaussians corresponds to the batch size, and select the batch sizes that yield the largest marginal likelihood, following the standard Bayesian model selection procedure. However, original B3O cannot apply to AL and constrained cases. Simple extension with constraining acquisition function or changing to AL acquisition function could apply to them but we do not investigate in this paper. B3O tends to select around 4-5 batch sizes regardless of the dimension, and is not applicable to large batch size. Moreover, Gaussian mixture model assumption is not always appropriate (e.g. Tanimoto kernel in drug discovery), whereas AdaBatAL naturally adopts these kernel via MMD.

\paragraph{Thompson sampling (TS)}
Thompson sampling (TS) \citep{hernandez2017parallel} is a random sampling method of $P(x^* \mid \textbf{D}_t)$ by maximising the function samples drawing from the predictive posterior. Due to its random sampling nature, exactly maximising the function samples is not strict when compared to others (e.g. hallucination). Thus, in practice, TS is typically done by taking argmax of function samples amongst the candidates of random samples over input space. This two-step sampling nature (random samples over input space $\rightarrow$ subsamples with argmax of random function samples) allows us for domain-agnostic BO. However, this scheme itself is a type of acquisition function, so other acquisition function is not naïvely supported. Moreover, due to the random sampling nature, the selected batch samples are not sparsified to efficiently explore uncertain regions.

\paragraph{Hallucination}
Hallucination \citep{azimi2010batch} tackled batch BO by simulating a sequential process by putting `fantasy' oracles estimated by GP, translating batch selection into a sequential problem.  Hallucination is successful in low batch size $n$, but not scalable. Even a single iteration of acquisition function maximisation is not trivial due to non-convexity, but they repeat this over $n$ times and produce prohibitive overhead. For discrete and mixed space, maximizing the acquisition function requires enumerating all possible candidates. However, the higher the dimension and larger the number of categorical classes, the more infeasibly large the combination becomes  (combinatorial explosion).

\paragraph{Local penalisation (LP)}
Local penalisation \citep{gonzalez2016batch}, simulates only acquisition function shape change, without fantasy oracles, by penalising acquisition function assuming Lipschitz continuity. This succeeds in speeding up the hallucination algorithm. However, the principled limitations are inherited (combinatorial explosion). Large batch sizes are also not applicable because maximising acquisition function still produces large overhead. This is because maximising acquisition function is typically computed by a multi-start optimiser, but the number of random seeds needs to increase dependent on the number of dimensions and multimodality of the true function. This optimiser also does not guarantee to be globally maximised, which contradicts the assumption of acquisition function (only optimal if it is globally maximised.). Furthermore, Lipschitz continuity assumption limits its applicable range to be only for continuous space. 

\paragraph{TurBO}
TurBO \citep{eriksson2019scalable} introduced multiple local BO bounded with trust regions, and allocates batching budgets based on TS. This succeeded in scalable batching via maintaining local BOs that are compact, via shrinking trust regions, based on heuristics with many hyperparameters. Selecting hyperpameters is non-trivial and TurBO cannot apply to discrete and non-Euclidean space, for which kernels do not have lengthscale hyperparameters for the trust region update heuristic (e.g. Tanimoto kernel for drug discovery \citep{ralaivola2005graph}).

\paragraph{SOBER}
SOBER \citep{adachi2022fast} first introduced the idea of batch BO as a kernel quadrature. Our AdaBatAL is based on SOBER when applying to batch BO tasks. The details are delineated in the section~\ref{sup:sober}. However, original SOBER is not capable of adaptive batch size or constrained cases.

\subsubsection{Constrained Batch Bayesian Optimization}
\paragraph{Constrained Expected Improvement (cEI)}
Constrained expected improvement (cEI) \citep{benjamin2019constraned} is the method based on constrained expected improvement acquisition function \citep{jones1998efficient}. cEI simply multiplies the probability of constraint satisfaction $q_\ell$ to the acquisition function. We adopted the official implementation on BoTorch \citep{balandat2020botorch}. The batching algorithm is based on sample average approximation, a standard batching methid in BoTorch library \citep{balandat2020botorch}.

\paragraph{Predictive Entropy Search with Constraints (PESC)}
Predictive Entropy Seach with Constraints (PESC) \citep{hernandez2015predictive} is the constrained version of the predictive entropy search acquisition function \citep{hernandez2014predictive}. The official implementation in Spearmint is dependent on Python 2 and is no longer supported in 2023. Thus, we adopted the implementation on BoTorch \citep{balandat2020botorch}. The batching algorithm is based on Monte Carlo sampling following the original code. However, this code is tremendously slow, which is repeatedly pointed out in BO literature \citep{eriksson2021scalable}. We set 7 days as the practical limit of execution time allowing for active learning, and PESC exceeds this limit for almost all tasks except for Hartmann synthetic function. Thus, we only compare PESC on Hartmann task but it was not the best performer.

\paragraph{Scalable Constrained Bayesian Optimization}
Scalable Constrained Bayesian Optimization (SCBO) is the constrained version of TurBO based on the TS acquistion function and trust region methods. We adopted the official implementation on BoTorch \citep{balandat2020botorch} and the same hyperparameters in the original papers \citep{eriksson2019scalable} for trust region update heuristics. 

\paragraph{Constrained Thompson sampling (cTS)}
Constrained Thompson sampling (cTS) is the constrained TS method. cTS has not been considered in existing work but this is a simple modification of SCBO. We adopted the two-step sampling used in SCBO for TS and removed the trust region heuristics because this cannot apply to a non-Euclidean kernel (e.g. Tanimoto kernel does not have lengthscale hyperparameter). This is coded based on SCBO implementation on BoTorch \citep{balandat2020botorch}.

\paragraph{PropertyDAG}
PropertyDAG \cite{park2022propertydag} is the method based on qNEHVI acquisition function and \citep{daulton2020differentiable, daulton2021parallel} for multi-objective optimization. This method assumes (1) ordered constraints but the constraint function is given, (2) multi-objective BO. So it cannot simply apply to our setting as it is. This method is the only one considering ordered case, so we dismantle the components of PropertyDAG to compare in the blackbox ordered constraint case. PropertyDAG consists of three parts: (A) explicit modelling of DAG network in surrogate model \citep{astudillo2021bayesian}, (B) zero inflation model to encode ordered constraint information to qNEHVI acquisition function, and (C) resampling of posterior function samples using sample average approximation to be more likely to satisfy the constraint. 
We cannot apply (A) and (B) for black-box ordered constraint, because (A) is only for white-box ordered constraint (we cannot model of unknown DAG), and (B) is only for multi-objective BO and specific acquisition function. Thus, we extracted the last part, (C) resampling with sample average approximation, and combined this with cEI, which we refer to PropertyDAG in this paper. We can say this as just resampled version of cEI. The implementation is based on cEI implementation on BoTorch \citep{balandat2020botorch} and added the resampling part.

\subsubsection{Batch Active Learning}
\paragraph{Maximum entropy (MaxEnt)}
Maximum entropy (MaxEnt) \citep{mackay1992information} is the classic acquisition function to select the next query with the largest Shannon entropy. As \citet{riis2022bayesian} pointed out, MaxEnt in FBGP is proportional to the posterior predictive variance. We adopted the following formulation \citep{riis2022bayesian}:
\begin{align}
    \text{MaxEnt} := \mathbb{H} \left[\int \mathbb{P}(y \mid x, \theta) \text{d} \mathbb{P}(\theta \mid \mathcal{D}_0) \right] \propto \mathbb{E}_{\mathbb{P}(\theta \mid \mathcal{D}_0)} \left[ C(x,x \mid \theta) \right]
\end{align}
For the batch construction, we take the top $n$ samples following the common practice in batch AL community \citep{kirsch2019batchbald}.

\paragraph{Bayesian Active Learning by Disagreement (BALD)}
Bayesian active learning by disagreement (BALD) \citep{houlsby2011bayesian} is another popular objective in Bayesian active learning, is to maximize the expected decrease in posterior entropy \citep{guestrin2005near}. \citet{houlsby2011bayesian} recast the objective from computing entropies in the parameter space to the output space by observing that it is equivalent to maximizing the conditional mutual information between the model’s parameters $\theta$ and output $\mathbb{I}[\theta, y \mid x, \mathcal{D}_0]$:
\begin{align}
    \text{BALD} := \mathbb{H} \left[ \mathbb{E}_{\mathbb{P}(\theta \mid \mathcal{D}_0)} \left[ y \mid x, \mathcal{D}_0, \theta \right] \right] - \mathbb{E}_{\mathbb{P}(\theta \mid \mathcal{D}_0)} \left[ \mathbb{H} \left[ y \mid x, \theta \right] \right]
\end{align}
\citet{kirsch2019batchbald} pointed out the original BALD criterion is independent selection of a batch of data points leads to data inefficiency as correlations between data points in an acquisition batch are not taken into account. Instead, BatchBALD is proposed whereby we jointly score points by estimating the mutual information between a joint of multiple data points and the model parameters:
\begin{align}
    \text{batchBALD} := \mathbb{H} \left[ \mathbb{E}_{\mathbb{P}(\theta \mid \mathcal{D}_0)} \left[ y_1,\dotsc,y_n \mid x_1,\dotsc,x_n, \mathcal{D}_0, \theta \right] \right] - \mathbb{E}_{\mathbb{P}(\theta \mid \mathcal{D}_0)} \left[ \mathbb{H} \left[ y_1,\dotsc,y_n \mid x_1,\dotsc,x_n, \theta \right] \right]
\end{align}
We adopted BatchBALD formulation for batch construction.

\paragraph{Bayesian Query-by-Committee (B-QBC)}
\citet{richardson2017gaussian} propose a Bayesian version of the Query-by-Committee \citep{seung1992query}, using the MCMC samples of the hyperparameters’ joint posterior.
We query a new data point where the mean predictions $m(x \mid \theta)$ disagree the most. Each mean predictor $m(\cdot \mid \theta)$ drawn from the posterior is equivalent to a single model, and thus this criteria can be seen as a Bayesian variant of a Query-by-Committee, and thus denoted as Bayesian Query-by-Committee (B-QBC). Given that $\bar{m}(x)$ is the average mean function, B-QBC is given as:
\begin{align}
    \text{B-QBC} := \mathbb{V}_{\mathbb{P}(\theta \mid \mathcal{D}_0)} \left[ m(x \mid \theta) \right] = \mathbb{E}_{\mathbb{P}(\theta \mid \mathcal{D}_0)} \left[ \left( m(x \mid \theta) - \bar{m}(x) \right)^2 \right]
\end{align}
For the batch construction, we take the top $n$ samples following the common practice in batch AL community \citep{kirsch2019batchbald}.

\paragraph{Active Bayesian CoreSets with Frank-Wolfe optimization (ACS-FW)}
Active Bayesian CoreSets with Frank-Wolfe optimization (ACS-FW) recasts the batch construction as the Bayesian coreset task. Our AdaBatAL is based on ACS-FW when applying to batch AL tasks. The details are deliniated in the section~\ref{sup:acs}. However, original ACS-FW is not capable of adaptive batch size nor constrained cases. Also, the Bayesian coreset formulation fails to incorporate the predictive uncertainty for batch construction unlike the kernel quadrature formulation. We implemented ACS-FW via following the official code \href{https://github.com/rpinsler/active-bayesian-coresets}{https://github.com/rpinsler/active-bayesian-coresets}.

\subsection{Dataset}
All datasets and tasks are summarized in Table~\ref{tab:tasks}.

\begin{table}[hbt!]
\centering
\caption{Summary of tasks.}
\label{tab:tasks}
\begin{tabular}{llllllll}
\toprule
 task & method & real/synthetic & space $\mathcal{X}$ & dimension & constraints & batch size & kernel\\
\midrule
 Hartmann & BO & synthetic & continuous & 6 & - & 5-90 & RBF\\
 Branin & cBO & synthetic & continuous & 2 & 2 & 20 & RBF\\
 \midrule
 Hartmann & AL & synthetic & continuous & 6 & - & 10 & RBF\\
 Ishigami & AL & synthetic & continuous & 3 & - & 10 & RBF\\
 Friedman & AL & synthetic & continuous & 5 & - & 10 & RBF\\
 Electrolyte & cAL & real-world & continuous & 3 & 2 & 10 & RBF\\
 Cantilever & cAL & real-world & continuous & 4 & 2 & 10 & RBF\\
 Steel & cAL & real-world & continuous & 9 & 1 & 10 & RBF\\
 \midrule
 Ackley & cBO & synthetic & mixed & 23 & 2 & 200 & RBF\\
 Hartmann & cBO & synthetic & continuous & 6 & 2 & 5 & RBF\\
 PestControl & cBO & real-world & discrete & 15 & 2 & 200 & RBF\\
 Malaria & cBO & real-world & discrete & molecule & 4 & 100 & Tanimoto\\
 FindFixer & cBO & real-world & graph & node & 3 & 100 & graph diffusion\\
 TeamOpt & cBO & real-world & graph & subgraph & 3 & 100 & graph diffusion\\
\bottomrule
\end{tabular}
\end{table}

\subsubsection{Synthetic Functions}
\paragraph{Hartmann}
Hartmann 6-dimensional function is defined as:
\begin{align}
f(x) &:= -\sum_{i=1}^4 \alpha_i \exp \left( - \sum_{j=1}^6 A_{ij} (x_j - P_{ij})^2 \right),\\
\alpha &= (1.0, 1.2, 3.0, 3.2)^\top, \\
\textbf{A} &= \begin{pmatrix}
10 & 3 & 17 & 3.5 & 1.7 & 8\\
0.05 & 10 & 17 & 0.1 & 8 & 14\\
3 & 3.5 & 1.7 & 10 & 17 & 8\\
17 & 8 & 0.05 & 10 & 0.1 & 14\\
\end{pmatrix}, \\
\textbf{P} &= \begin{pmatrix}
1312 & 1696 & 5569 & 124 & 8283 & 5886\\
2329 & 4135 & 8307 & 3736 & 1004 & 9991\\
2348 & 1451 & 3522 & 2883 & 3047 & 6650\\
4047 & 8828 & 8732 & 5743 & 1091 & 381\\
\end{pmatrix}
\end{align}
We take the negative Hartmann function as the objective of BO to make this optimisation problem maximisation. All input variables are continuous with bounds $[0, 1]^6$. The batch size $n$ is 100. The continuous prior is the uniform distribution ranging from [0, 1], following \cite{adachi2023sober}. The noisy output is generated by adding i.i.d. zero-mean Gaussian noise with the $0.0192^2$ variance to the noiseless $f(x)$.

For constrained BO, we added two constraints; (1) $\sum_{i=1}^d x_i \geq 0.15$ and (2) $\sum_{i=1}^d x_i \leq 3$.

\paragraph{Branin}
Branin function is defined as:
\begin{align}
    f(x) := \prod_{i=1}^d \frac{\sqrt{\sin(x) + 0.5 \cos(3x)}}{\sqrt{0.5x} + 0.3},
\end{align}
where the dimension $d=2$. All input variables  are continuous with bounds $x \in [-2, 3]^d$. The batch size $n$ is 20. The continuous prior is the uniform distribution. The noisy output is generated by adding i.i.d. zero-mean Gaussian noise with the $0.0192^2$ variance to the noiseless $f(x)$.

For constrained BO, we added two constraints; (1) $\sum_{i=1}^d x_i^2 \leq 4$ and (2) $\sum_{i=1}^d x_i \leq 0$.

\paragraph{Ishigami}
Ishigami function is defined as:
\begin{align}
    f(x) := \sin(x_1) + 7 \sin^2(x_2) + 0.1 x_3^4 \sin(x_1),
\end{align}
where $x_i$ is the $i$-th dimensional input and the dimension $d=3$. All input variables are continuous with bounds $x \in [-\pi, \pi]^d$. The batch size $n$ is 10. The continuous prior is the uniform distribution. The noisy output is generated by adding i.i.d. zero-mean Gaussian noise with the $0.187^2$ variance to the noiseless $f(x)$.

\paragraph{Friedman}
Friedman function is defined as:
\begin{align}
    f(x) := 10\sin(\pi x_1 x_2) + 20 (x_3 - 0.5)^2 + 10 x_4 + 5x_5,
\end{align}
where $x_i$ is the $i$-th dimensional input and the dimension $d=5$. All input variables are continuous with bounds $x \in [0, 1]^d$. The batch size $n$ is 10. The continuous prior is the uniform distribution. The noisy output is generated by adding i.i.d. zero-mean Gaussian noise with the $0.05^2$ variance to the noiseless $f(x)$.

\paragraph{Ackely}
Ackley funciton is defined as:
\begin{align}
f(x) := - a \exp \left[ -b \sqrt{\frac{1}{d} \sum_{i=1}^d x_i^2} \right] - \exp \left[
\frac{1}{d} \sum_{i=1}^d \cos (c x_i) \right] + a + \exp(1)
\end{align}
where $a = 20, c = 2\pi, d = 23$. We take the negative Ackley function as the objective of BO to make this optimisation problem maximisation. We modified the original Ackley function into a 23-dimensional function with the mixed spaces of 3 continuous and 20 binary inputs from $[0, 1]^{20}$, following \cite{adachi2023sober}. The batch size $n$ is 200. The continuous prior is the uniform distribution ranging from [-1, 1]. The binary prior is the Bernoulli distribution with unbiased weights of 0.5. We assume each of the continuous and binary priors at each dimension is independent.

For constrained BO, we added the two constraints; (1) $x_1 \geq 0$ and (2) $x_2 \geq 0$, where $x_1$ and $x_2$ are the first and second dimensions of continuous inputs.

\subsubsection{Real-World Functions}
\paragraph{Electrolyte}
Electrolyte is the new problem for the AL task. This is the task of creating the model that predicts the ionic conductivity for the given composition of liquid electrolyte material for the next generation of lithium-ion batteries. This ionic-conductivity function is used for the control model of batteries and plays a crucial role in the control accuracy. However, common practice is to use the lookup table with massive data or pairwise linear function fitting. Collecting the ionic conductivity data requires costly laboratory experiments and fewer data points can accelerate this process while minimizing the cost. GP and AL are powerful frameworks to offer more accurate models with fewer data sizes and cheap models allowing them to be implemented in the control chip. Still, this data collection is under an unknown constraint; the freezing point. While low-temperature operation performance is the key performance indicator of batteries, it causes freezing electrolytes and cannot measure ionic conductivity. The freezing point is dependent on both lithium salt molarity and the cosolvent composition. They show the complex non-linear relationship due to the solvation effect and cannot predict even with the state-of-the-art quantum chemistry simulator. Thus, it is natural to assume this freezing point is an unknown constraint. We create the true function by fitting the experimental data of MA-DMC-EMC-LiPF$_6$ \citep{logan2018study} system using the Casteel-Amis equation \citep{casteel1972specific}. Note that Casteel-Amis equation is just for the interpolation of experimental data to be continuous, and is not capable of predicting different cosolvent nor freezing points. 

Electrolyte is a three-dimensional continuous input function with two constraints. The input features are (1) the lithium salt (LiPF$_6$) molarity, (2) DMC/EMC cosolvent ratio, and (3) MA/carbonates cosolvent ratio, respectively. The inputs are bounded with $x_1 \in [0, 2]$, $x_2 \in [0, 1]$, and $x_3 \in [0, 0.3]$. The constraints are $x_1 > 0.3$ and $x_2 < 0.9$. The noisy output is generated by adding i.i.d. zero-mean Gaussian noise with the $3^2$ variance to the noiseless $f(x)$.

\paragraph{Cantilever}
Cantilever \citep{wu2001safety} has been proposed for a task to develop a probability-based design optimization framework for ensuring high reliability and  safety. This task is to design a cantilever beam under the two failure modes as safety constraints. The objective function to model with GP is the tip displacement, modelled as:
\begin{align}
    &f(x) := \frac{4 \times 100^3}{E} \sqrt{X^2 + Y^2},\\
    &\text{subject to:}\\
    &\frac{f(x) - 4400}{3100} < 2.2535,\\
    &0.8 (X + Y) < R.
\end{align}

Cantilever is a four-dimensional continuous input function with two constraints. The input features are (1) the yield stress $R$, (2) the Young's modulus of beam material $E$, (3) the horizontal load $X$, and (4) the vertical load $Y$, respectively. The inputs are bounded with $R \in [3E+4, 5E+4]$, $E \in [1E+7, 5E+7]$, $X \in [1E+2, 1E+3]$, and $Y \in [5E+3, 5E+4]$. The noisy output is generated by adding i.i.d. zero-mean Gaussian noise with the $1^2$ variance to the noiseless $f(x)$.

\paragraph{Steel}
Steel \citep{kuschel1997two} has been proposed for design optimization to balance the reliability and cost. This task is to design a steel column under cost constraints. The objective function to model with GP is the limit state function, modelled as:
\begin{align}
    &f(x) := F_s - P \left[ \frac{1}{2 B D} + \frac{F_0 E_b}{B D H (E_b - P)}\right],\\
    &\text{subject to:}\\
    &B D + 5H < 9000,
\end{align}
where
\begin{align}
    P &:= P_1 + P_2 + P_3\\
    E_b &:= \frac{\pi^2 E B D H^2}{2 L^2}
\end{align}
Steel is the nine-dimensional continuous input function with one constraint. The input features are (1) the yield stress $F_s$, (2) the dead weight load $P_1$, (3) the variable load $P_2$, (4) the variable load $P_3$, (5) the flange breadth $B$, (6) the flange thickness $D$, (7) the profile height $H$, (8) the initial deflection $F_0$, and (9) Young's modulus $E$, respectively. The inputs are bounded with $F_s \in [300, 500]$, $P_1 \in [1E+4, 1E+5]$, $P_2 \in [4E+5, 1E+6]$, $P_3 \in [4E+5, 1E+6]$, $B \in [290, 310]$, $D \in [14, 26]$, $H \in [290, 310]$, $F_0 \in 209800, 210100]$. The noisy output is generated by adding i.i.d. zero-mean Gaussian noise with the $1^2$ variance to the noiseless $f(x)$.

\paragraph{PestControl}
Pest Control (PestControl in the main) is proposed in \citet{oh2019combinatorial}, which is a multi-categorical optimisation problem (15 dimensions, 5 categories for each dimension). We wish to optimise the effectiveness of pesticides by choosing the 5 actions (selection of pesticides from 4 different firms, or not using any of them), but penalised by their prices. This choice is a sequential decision of 15 stages, and the objective function is expressed as the cumulative loss function with the total of both cost and the portion having pest. The batch size $n$ is 200. We set the categorical prior with equal weights for each choice (discrete uniform distribution). Code is used in \href{https://github.com/xingchenwan/Casmopolitan}{https://github.com/xingchenwan/Casmopolitan} \citep{wan2021think}.

We added 2 constraints for a more realistic situation. 
The first constraint is ecosystem change, which assumes exterminating pests too much causes other harmful pests/animals to increase when they reach the hidden threshold. The portion of the product having pests follows the dynamics below:
\begin{align}
    z_i &= \alpha_i (1 - x_i) (1 - z_{i-1}) + (1 - \Gamma_i x_i) z_{i-1},\\
    z_i &\geq z_\text{limit}, \label{eq:ctpest}
\end{align}
where $i$ is the number of pest control cycles (15 in total), $z$ is the portion of the product having pest, $x$ is the effectiveness of pesticide that follows a beta distribution with the parameters, which has been adjusted according to the sequence of actions taken in previous control points, $\alpha$ is the action taken (selection of pesticides from 4 different firms, or not using any of it), and $z_\text{limit}$ is the threshold for ecosystem change (we set $1e-3$). Eq. \ref{eq:ctpest} is the constraint of ecosystem change, and we assume the latent variable $z_i$ is observable.

The second constraint is neighbour disputes, which assume some of the pesticides have unfavourable smells.
Neighbours objection follows the Bernoulli distribution and its weights based on the proportion of certain pesticide types and random Gaussian noise. Thus, the feedback to this constraint is in the noisy binary value. If the neighbours' objection is larger than supportive opinion $\theta_\text{pest} > 0.5$, a decision maker stops spraying pesticides, thus, objective value cannot be evaluated.

\paragraph{Malaria}
The objective is to discover an anti-malarial drug exhibiting the smallest EC50 value, which is defined as the concentration of the drug that gives half the maximal response. The lower the concentration, the more effective (better) the drug. The dataset consists of 20,746 small molecules taken from the P. falciparum whole-cell screening derived by the Novatis-GNF Malaria Box \citep{spangenberg2013open}. The molecules are represented as SMILES string and are converted into 2048-dimensional binary features for the Tanimoto kernel.
We set four safety constraints, all of which are rules of thumb for judging molecules likely to be oral drugs, shared in drug discovery community \citep{lipinski1997experimental, veber2002molecular, butler2004role, mochizuki2019qex}.

The first is Lipinski's rule of five \citep{lipinski1997experimental}, (A) no more than 5 hydrogen bond donors, (B) no more than 10 hydrogen bond acceptors, (C) A molecular mass less than 500 daltons, (D) A calculated octanol-water partition coefficient that does not exceed 5, (E) no more than 5 rotatable bonds.
The second is the Veber filter \citep{veber2002molecular}, (A) no more than 10 rotatable bonds, (B) a polar surface area that does not exceed 140.
The third is the REOS filter \citep{butler2004role}, (A) A molecular mass more than 200 daltons and less than 500 daltons, (B) A calculated octanol-water partition coefficient that exceeds -5 but does not exceed 5, (C)  no more than 5 hydrogen bond donors, (D) no more than 10 hydrogen bond acceptors, (E) no more than 8 roratable bonds, (F) more than 15 but less than 50 heavy atoms, (G) more than -2 but less than 2 formal charge.
The fourth is the drug likeliness filter, (A) A molecular mass less than 400 daltons, (B) at least one ring structure, (C) no more than 5 roratable bonds, (D) no more than 5 hydrogen bond donors, (E) no more than 10 hydrogen bond acceptors, (F) A calculated octanol-water partition coefficient that does not exceed 5.

\paragraph{FindFixer}
This task is to find the fixer connecting influencers rather than finding the most popular influencer on the social networks graph. A job seeker who wishes to be a celebrity explores the fixer to ask introductions based on graph data using the centrality analysis. Finding a node requires searching on a website or meeting in person, both of which are expensive to evaluate. Fixer can be interpreted as a node with maximum eigenvector centrality under constraints on the degree centrality that does not exceed the threshold \citep{kiss2008identification}. In other words, finding the node that is connected to the largest number of nodes with many edges but does not have many edges itself. A job seeker wishes to find the fixer who connects influencers with similar popularity (degree centrality).  Thus, the node is constrained based on the degree centrality, and other hidden preference factors. A job seeker judges constraints as a binary value, and the judgment is possibly shaky. We assume the domain is defined as a social network graph synthesized by the Barábsi–Albert model (BA) \citep{barabasi1999emergence}.

\paragraph{TeamOpt}
This task is to organise a team consisting of the most diverse skill sets of members \citep{wan2023bayesian}. The objective is measured by the entropy of the skills of members, assuming the optimal team is when each member is specialised in one skill, and the whole skill distribution is close to uniform. Such teams are positioned on the node of the supergraph, of which edge is the similarity between teams defined as the Jaccord index. 
The constraints are interpersonal relationships. Every combination of two individuals from $N$ candidates has unobservable hidden continuous likability from 0 to 1. The first is the mean likability constraint, which is the mean of likeability between all possible combinations of members that should be larger than equal-chance. The second is the tragedy-avoidance constraint, which is a binary judge that none of them has a likability lower than a threshold. The third is a flat-relationship constraint, which assumes an entropy of likability must be higher than a threshold. As likability is unobservable, a decision-maker needs to seek advice from many colleagues who partially know each constraint but are noisy estimations.

\subsection{Complexity Analysis}
\begin{figure}[ht!]
  \centering
  \includegraphics[width=0.4\hsize]{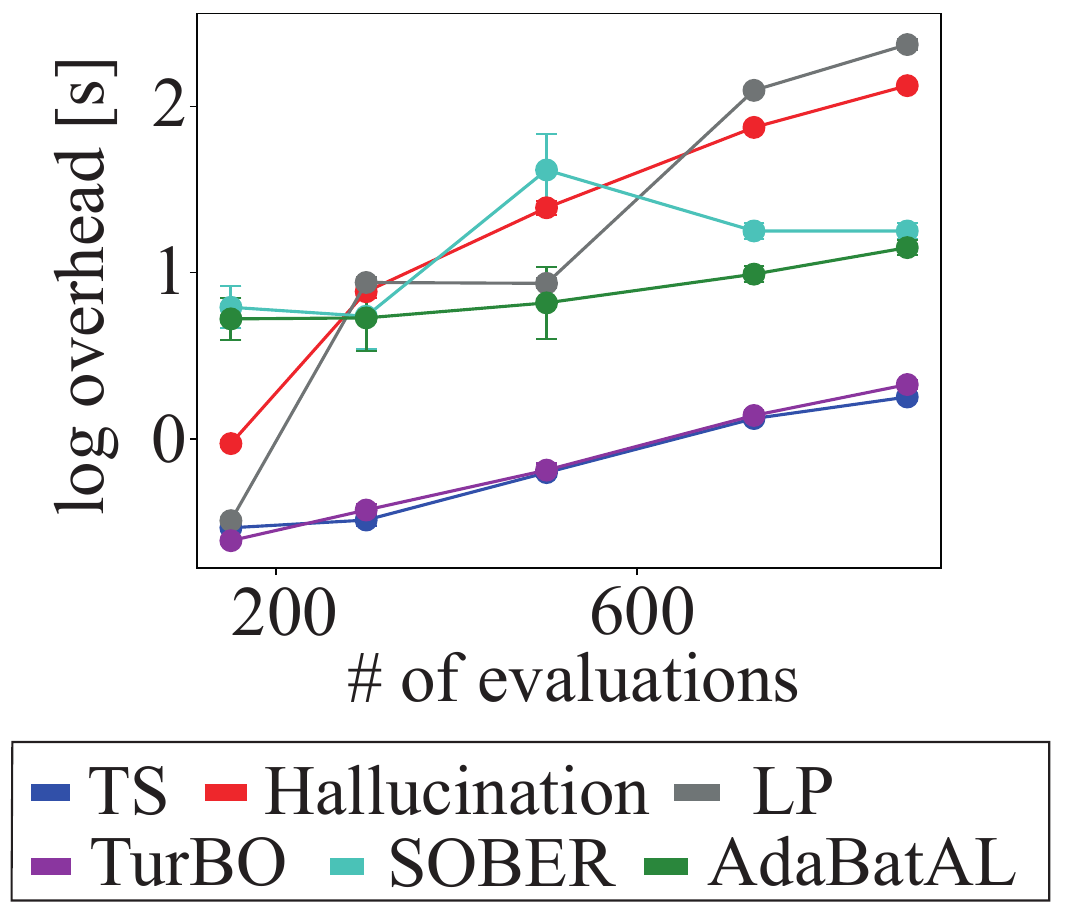}
  \caption{Overhead on Batch Bayesian optimization task for Hartmann ($d=6$)}
  \label{fig:overhead}
  \vspace{-1em}
\end{figure}
As explained in the section~\ref{sec:complexity}, the time complexity of the AdaBatAL is lower than $\mathcal{O}(N M + M^2 \log n + M n^2 \log (N / n))$ \citep{hayakawa2022positively}, where $N$ is the number of unlabelled pool, $M$ is the number of Nyström samples, and $n$ is the upper bound of the batch size. The space complexity is $\mathcal{O}(N M)$.

We empirically compare the time complexity against the baselines using the Hartmann function with unconstrained batch BO tasks. Figure~\ref{fig:overhead} shows the log overhead to generate the batch samples with different batch sizes that are the same with Figure~\ref{fig:batchsize} setting. While TurBO and TS were faster than others, our AdaBatAL was relatively faster than other baselines (SOBER, hallucination, and LP).

\end{document}